\documentclass[mathpazo]{csam}

\usepackage{caption}
\usepackage{booktabs,multirow}
\usepackage{tabularx}
\usepackage{multirow}
\usepackage{array}
\usepackage{arydshln}
\newcolumntype{Y}{>{\centering\arraybackslash}m} 

\usepackage{cite}
\usepackage[numbers,sort&compress]{natbib}
\usepackage{epstopdf}
\usepackage{color}
\usepackage{slashed}
\usepackage{graphicx} 
\usepackage{float} 
\usepackage{subfigure} 
\usepackage{algorithmic}
\usepackage{algorithm}
\usepackage{lipsum}



\begin{document}

\title{Review of Mathematical Optimization in Federated Learning}

\author[S. Yang et~al.]{Shusen Yang\affil{1,2},~%\comma,%\corrauth,
      Fangyuan Zhao\affil{2},~Zihao Zhou\affil{1}, Liang Shi\affil{2},\\~Xuebin Ren\affil{2}, and Zongben Xu \affil{1,2}}
\address{\affilnum{1}\ School of Mathematics and Statistics,
         Xi'an Jiaotong University,
         China. \\
          \affilnum{2}\ National Engineering Laboratory for Big Data Analytics,
          Xi'an Jiaotong University,
         China.}
\emails{{\tt shusenyang@mail.xjtu.edu.cn} (S.~Yang), {\tt chan@email} (F.~Zhao),
         {\tt zhao@email} (Z.~Zhou)}

\begin{abstract}
{
Federated Learning (FL) has been becoming a popular interdisciplinary research area in both applied mathematics and information sciences.
Mathematically, FL aims to collaboratively optimize aggregate objective functions over distributed datasets while satisfying a variety of privacy and system constraints. 
Different from conventional distributed optimization methods, FL needs to address several specific issues (e.g., non-i.i.d. data distributions and differential private noises), which pose a set of new challenges in the problem formulation, algorithm design, and convergence analysis.
In this paper, we will systematically review existing FL optimization research including their assumptions, formulations, methods, and theoretical results. 
Potential future directions are also discussed.

}

\end{abstract}

\ams{90C26, 90C31, 68W15, 68T05
%The information of the AMS subject classification can be found in http://mathscinet.ams.org/msc/msc2010.html
}

%%%% maketitle %%%%%
\maketitle

%%%% Start %%%%%%
\section{Introduction}
\label{sec1}

{
With the increasingly stringent privacy regulations~\cite{voigt2017eu, pardau2018california}, data isolation has been becoming the key bottleneck of data sciences and artificial intelligence. To address this issue, Federated learning (FL) emerges as a popular privacy-preserving distributed Machine Learning (ML) paradigm, which enables multiple data owners to jointly train ML models without sharing the raw data~\cite{kalra2023decentralized,hard2018federated,kairouz2021advances}. It has gained extensive interests from both academia and industry, and demonstrated great success across multiple domains, including medicine~\cite{ogier2023federated}, finance~\cite{yang2019ffd}, and industry~\cite{zhang2024vertical}, etc.

Mathematically, FL training tasks are essentially distributed optimization problems, which aim to minimize aggregate global objectives (e.g., the mean empirical loss), across a set of distributed data owners by exchanging model parameters trained on their local datasets~\cite{zhang2023understanding, touri2023unified, zhao2024faster}. Despite being similar in essence, FL optimization has many distinct characteristics from traditional distributed optimization. Their main difference lies in the communication environment. Specifically, traditional distributed optimization, mainly in the form of distributed ML~\cite{beznosikov2023biased,doi:10.1137/17M1157891,doi:10.1137/18M1194699} (some used in distributed resource allocation~\cite{doi:10.1137/19M127879X,doi:10.1137/21M1400924,doi:10.1137/17M1151973}), is often used for high-throughput ML speedup in data centers. Here, multiple homogeneous computing nodes with uniformly distributed data partitions are connected by reliable networks with Gigabytes of bandwidth. However, FL is often applied to achieve collaborative and privacy-preserving ML in wide-area networks, where geographically distributed clients with naturally generated data collaborate to train ML models over bandwidth-constrained communication channels.

Due to the drastic differences, FL needs to address a variety of specific and complicated issues in optimization. For example, naturally generated data at the FL clients are commonly heterogeneous, i.e., non-balanced and non-i.i.d. distributed, which leads to biased model aggregation. Considering limited communication bandwidth, FL often performs multiple local updates before the global aggregation, further amplifying the model bias caused by data heterogeneity.
Furthermore, model dissemination and aggregation give rise to concerns of private information leakage and falsification. The decentralized communication architecture also leads to partially local approximations of global objective functions.
These issues pose new challenges in optimization in FL, including
problem formulation, algorithm design, and convergence analysis. In particular, typical challenges
are summarized as follows:

\begin{itemize}

\item \textbf{Biased local objectives from non-i.i.d. datasets.} 
The potentially unknown and non-i.i.d. data distributions among distributed nodes\footnote{The terms ''node'', ''client'' and ''data owner'' are used interchangeably within this survey.} could result in biased local objectives~\cite{zhu2021federated, kairouz2021advances}. Thus, the global objective function in FL optimization cannot be decomposed into trainable local objectives without bias. This means that the gradients of local models may largely deviate from the steepest descent direction of the global objective, leading to significant degradation of convergence speed and model accuracy~\cite{karimireddy2020scaffold}.

\item \textbf{Perturbed gradients with DP noises.}
The Differential Privacy (DP) mechanisms~\cite{dwork2006calibrating} are commonly adopted in the FL optimization process to protect exchanged parameters or gradients, by introducing statistically unbiased noises (e.g., Gaussian~\cite{abadi2016deep} or Laplacian~\cite{wu2020value} noises). Despite the statistical unbiasedness, random noises may overwhelm the useful gradient information, causing unstable and slow convergence towards sub-optimal solutions~\cite{zhang2022understanding,hu2023federated, girgis2021shuffled}.

\item \textbf{Partially approximated objectives under decentralized topologies.}
In FL systems with decentralized typologies, the model exchange has to rely on limited peer-to-peer neighborhood communication. The global objective for each data owner is thus partially approximated by consolidating the local objectives of their adjacent neighbors~\cite{doi:10.1137/21M1465081}. Consequently, the aggregated gradient in each step is also a partial approximation to that of the global objective. The approximation error can lead to significantly slower convergence~\cite{yuan2016convergence, kalra2023decentralized}, especially in non-i.i.d. scenarios.

\item \textbf{Obsolete solutions in online settings.}
Online FL optimization aims to learn a series of global functions with minimized cumulative losses from distributed sequential data~\cite{hong2021communication,chen2020asynchronous,m2022personalized,kwon2023tighter}. However, the coming data may have a time-evolving distribution, resulting in inevitable generalization errors and instability of models on new data. Meanwhile, the inherent time-varying and cumulative constraints, e.g. communication, computing, and memory capabilities, may also complicate the problem modeling and solving~\cite{gauthier2022resource}. 
\end{itemize}

\begin{figure}[ht]
  \centering
  \includegraphics[scale=0.45, trim={0 170 0 10}, clip]{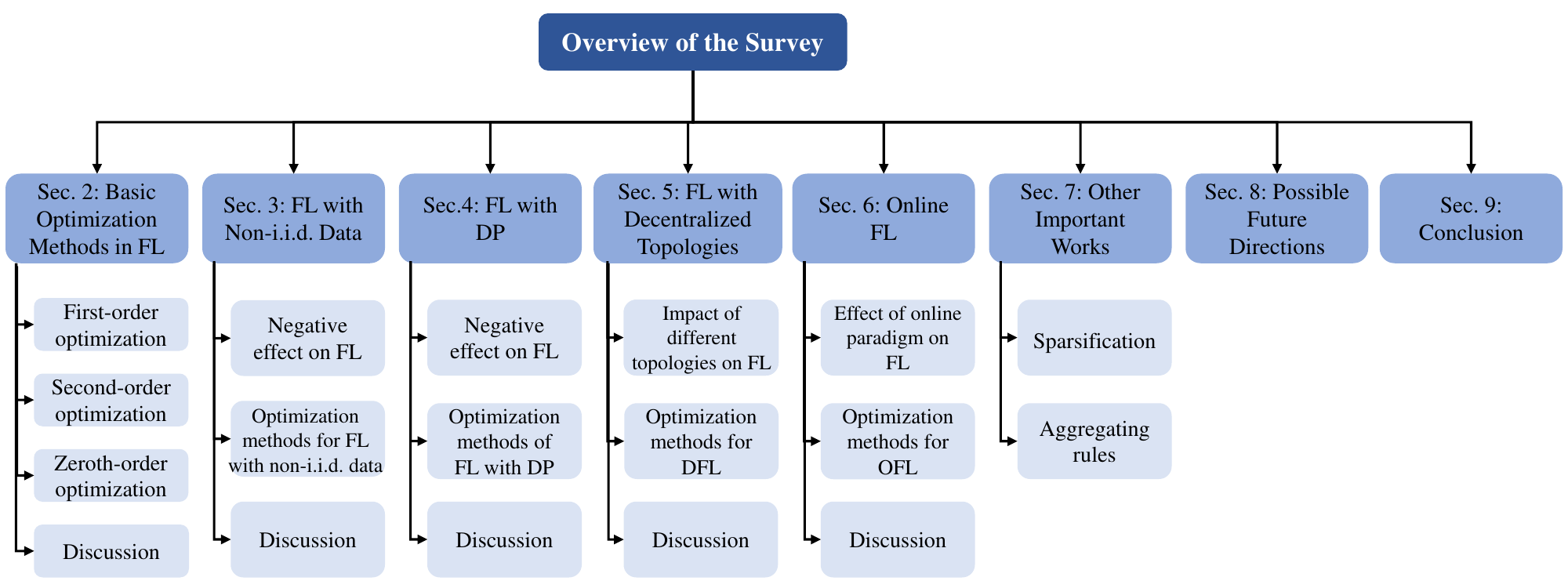}  
  \\
  \caption{The overview of the survey.}\label{fig:overview}
\end{figure}

This paper presents a systematic survey of the mathematical optimization in FL, summarizing the assumptions, problem formulations, optimization methods, and theoretical results.  
It is worth noting that although there exist several FL surveys~\cite{antunes2022federated,lu2024federated,kairouz2021advances, yang2019federated,ren2024belt, kumar2023impact}, all of them are presented from the perspective of information sciences, including the FL architectures~\cite{antunes2022federated}, algorithms~\cite{lu2024federated}, and applications~\cite{nguyen2022federated}, rather than mathematical optimizations like our survey. Fig.~\ref{fig:overview} illustrates the overview of this survey. We first review typical optimization methods in FL in Section~\ref{sec:optimizers}. 
The optimization challenges and solutions when dealing with non-i.i.d. data, DP noises, decentralized network topologies, and online FL optimizations are presented in Sections~\ref{sec:non-i.i.d.}-\ref{sec:online}, respectively. 
We then summarize other important works in Section~\ref{sec:other-works} and discuss possible future directions in Section~\ref{sec:future_direction}.
Finally, we conclude the survey in Section~\ref{sec:conclusion}.

\section{Basic optimization methods in Federated Learning}
\label{sec:optimizers}

The common FL optimization is mostly studied in the context of horizontal FL (HFL) where participants possess the same feature space but different samples~\cite{mcmahan2017communication,kairouz2021advances}. We will use HFL as the default FL setting \footnote{Another important category of FL is vertical FL (VFL), where participants share the same sample space but different features. More details can be referred to in \cite{liu2024vertical,jin2021cafe,wang2024unified}.} to introduce the general workflow of FL optimization.

\begin{figure}[htbp]
\centering 
\includegraphics[width=0.8\textwidth]{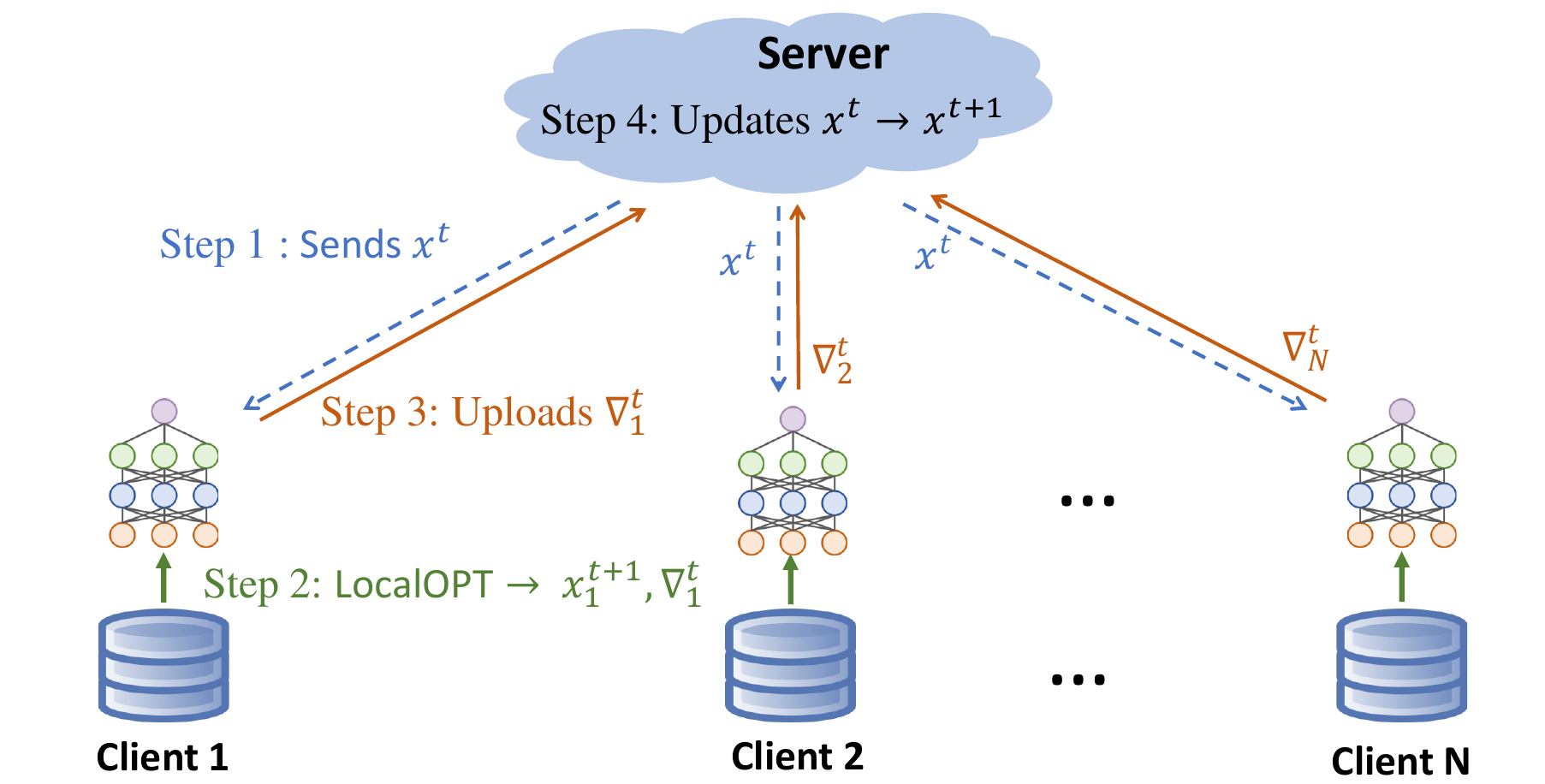} 
\caption{General framework and workflow of FL.}
\label{Fig:HFLworkflow} 
\end{figure}

\begin{algorithm}[t]
	\renewcommand{\algorithmicrequire}{\textbf{Input:}}
	\renewcommand{\algorithmicensure}{\textbf{Output:}}
	\caption{Algorithmic procedures of FL optimization}
	\label{alg1}
	\begin{algorithmic}[1]
		\REQUIRE Client number $N$, objective function $f(x) = \frac{1}{N} \sum_{i=1}^N  f_i(x)$, initialized $x_0$, total communication rounds $T$, local learning rate $\eta_{l}$, global learning rate $\eta_{g}$, number of local updates $E$ in one round, distributed dataset $\{\xi_1,\ldots,\xi_N\}$.
            \ENSURE $x^{T+1}$.  
            \FOR{$t \in\{1,\dots,T\} $}
            \STATE Server samples a subset $S^t$ of $K$ clients and server sends $x^t$ to these clients;
            \FOR{each client $i \in S^t$}
            \STATE $x_{i}^{t+1}$, $\nabla_i^t$=\textbf{LocalOPT}$\left(x^t, \eta_l,E, \xi_{i}\right)$;
            \ENDFOR
            \STATE Server aggregates $\nabla^t=\frac{1}{|\mathcal{S}^t|}\sum_{i\in\mathcal{S}^t}  \nabla_i^t$;
           \STATE Update global model $x^{t+1} =x^{t} + \eta_g \nabla^t$.
            \ENDFOR
	\end{algorithmic}  
\end{algorithm}

As shown in Algorithm~\ref{alg1} and Fig.~\ref{Fig:HFLworkflow}, a typical FL system consists of multiple distributed clients and one central server. At each iteration round $t$, the server randomly samples a subset $S^t$ from $N$ clients and broadcasts the current global model $x^t \in \mathbb{R}^d$ to these clients. Each client indexed by $i$ in $S^t$  then executes LocalOPT upon receiving $x^t$, which involves applying an optimization method to minimize their respective local objectives, subsequently obtaining a local model $x_i^{t+1}$ and uploading the local gradient $\nabla_i^t$ to the server. Once all gradients from $S^t$ are received, the server aggregates all local gradients to obtain the global gradient $\nabla^t$ and accordingly updates the global model as $x^{t+1}=x^t+\eta_g \nabla^t$, where $\eta_g$ is the global learning rate.

Given its pivotal role in local data information extraction, the choice of LocalOPT would significantly affect the performance of FL training process. Since LocalOPT essentially executes machine learning (ML) tasks in each round on the client side, its candidates encompass all ML optimization methods, including the first-order, second-order, and zeroth-order ones. Each type of these optimization methods exhibits unique convergence characteristics when being adapted to FL settings, which are summarized in Table~\ref{table:performance}. The following are common assumptions used in the convergence analysis of FL optimization algorithms~\cite{bubeck2015convex,reddi2020adaptive,zhao2024faster}.

\begin{itemize}
    \item \textbf{Lipschitz Objective Function (LOF):}
    $f(x)$ is $\beta$-Lipschitz continuous if there exists $\beta \geq 0$ such that for all $x_1, x_2 \in \mathbb{R}^d$,
    \begin{equation}
    |f(x_1)-f(x_2)|\leq \beta \|x_1-x_2\|.
    \end{equation}
    \item \textbf{Smooth Objective Function (SOF):}
    $f(x)$ is $L$-smooth if $f(x)$ has $L$-Lipschitz continuous gradient, i.e., for all $x_1, x_2 \in \mathbb{R}^d$, 
    \begin{equation}
    \|\nabla f(x_1)-\nabla f(x_2)\|\leq L\left\|x_1-x_2\right\|.
    \end{equation}
    When the second-order gradients of $f(x)$ is Lipschitz continuous, we call the condition \textbf{Lipschitz Hessian (LH)}. 
     \item \textbf{Strongly Convex Objective Function (SCOF):}
    $f(x)$ is $\mu$-strongly convex if there exists $\mu \geq 0$ such that for all $x_1,x_2\in \mathbb{R}^d$, 
    \begin{equation}
    f(x_1)\geq f(x_2)+(x_1-x_2)^T\nabla f(x_2)+\frac\mu2\|x_1-x_2\|_2^2.
    \end{equation}
     \item \textbf{Convex Objective Function (COF):}
    $f(x)$ is convex if for all $x_1,x_2\in \mathbb{R}^d$, it holds that
    \begin{equation}
    f(x_1)\geq f(x_2)+(x_1-x_2)^T\nabla f(x_2).
    \end{equation}
     \item \textbf{Coercive Function (CF):}
    $f(x)$ is coercive if $\lim_{\|x\| \to \infty} f(x)\rightarrow \infty$.
    \item \textbf{Bounded Gradient (BG):} The gradient of $f(x)$ is $G$-bounded if there exists $G \geq 0$ such that for all $x\in \mathbb{R}^d$,
    $\|\nabla f(x)\|\leq G$.
    \item \textbf{Bounded Variance (BV):}
    The variance of each stochastic gradient $\nabla f_i(x;\xi)$ is bounded if there exists $\sigma\in \mathbb{R}$, such that
    \begin{equation}
    \mathbb{E}_\xi\|\nabla f_i(x;\xi)-\nabla f_i(x)\|^2\leq\sigma^2,
    \end{equation}
    where $f_i(\cdot)$ denotes the local objective function of the $i$-th client, $x$ is the current model parameter and $\xi$ is the data sampled in the current round of local training.
     \item \textbf{Bounded Gradient Dissimilarity (BGD):}  Local gradients $\{\nabla f_{i} (x), \forall i\}$ satisfy $(G, B)$-bounded gradient dissimilarity if there exist constants $G \geq 0$ and $B \geq 1$ such that 
     \begin{equation}
         \frac{1}{N} \sum_{i=1}^{N} \| \nabla f_{i} (x) \|^2 \leq G^2 + B^2 \| \nabla f(x) \|^2, \quad \forall x\in \mathbb{R}^d.
     \end{equation}
\end{itemize}

In above assumptions, LOF, SOF, and LH describe the smoothness of the objective function. SCOF and COF characterize the convexity of objective functions. CF ensures that the objective function has a global minimum. BG, BV, and BGD capture the properties of gradients.

\begin{table}[ht]
\centering
\caption{Comparison of basic optimization methods in FL. }
\label{table:performance}
\begin{tabularx}{\textwidth}{Y{2.5cm}Y{5cm}Y{3cm}Y{2.8cm}}
\toprule
  \textbf{Categories}        & \textbf{Methods}                & \textbf{Assumptions}      & \textbf{Convergence rates}    
\\ 
\hline
  \multirow{8}{2.5cm}{\centering First-order  Methods}  & SGD & \multirow{2}{*}{SOF, BV, BG}           &  \multirow{2}{*}{$\mathcal{O}(\frac{\sigma}{\sqrt{KT}})$}            
  \\
  & (e.g.,\cite{stich2019error,yu2018parallel})                    &     &         
  \\  
  \cline{2-4}
        & {Momentum-based SGD} & \multirow{2}{*}{SOF, BV}           & \multirow{2}{*}{$\mathcal{O}(\frac{\sigma}{\sqrt{KT}})$}   \\
        & (e.g., \cite{karimireddy2021learning,yang2022federated}) & &
  \\
  \cline{2-4}
        & {Adam}   & \multirow{2}{*}{SOF, BV, BG} & \multirow{2}{*}{$\mathcal{O}(\frac{\sigma}{\sqrt{KT}})$}    \\
        & {(e.g., \cite{reddi2020adaptive,ju2023accelerating})} & &
  \\    
  \cline{2-4}
        & {ADMM}    & \multirow{2}{*}{SOF, CF}          & \multirow{2}{*}{$\mathcal{O} (\frac{1}{T})$}      \\
        & {(e.g., \cite{zhou2023federated,jeon2021privacy,kant2022federated})} & &
    \\     
\hline   
\multirow{2}{2.5cm}{\centering Second-order Methods} & \multirow{2}{3cm}{\centering Newton (e.g.,~\cite{elgabli2022fednew,safaryan2021fednl,dinh2022done,nagaraju2023fonn})}    & \multirow{2}{*}{SOF, LH, SCOF}   & \multirow{2}{*}{$\mathcal{O}(\gamma^{-2^T}) (\gamma>1)$} 
\\    
&  & &  
\\
\hline
\multirow{2}{2.5cm}{\centering Zeroth-order Methods} & \multirow{2}{2.5cm}{\centering ZO-SGD (e.g.,~\cite{fang2022communication,haddadpour2019local,qiu2023zeroth})}     &  \multirow{2}{*}{LOF}   & \multirow{2}{*}{$\mathcal{O} ( \sigma \sqrt{\frac{d}{KT}} )$}    
\\ 
&  & & 
\\
\bottomrule
\end{tabularx}
\end{table}

\subsection{First-order optimization methods}

First-order optimization methods~\cite{beck2017first} rely only on first-order gradients for the model updating in LocalOPT. They are widely adopted in FL due to their lower computational requirements for gradient estimation, especially in settings with pronounced computational heterogeneity. 
Here, we introduce several representative first-order optimization methods in FL.

The most common choice for the first-order method is Stochastic Gradient Descent (SGD)~\cite{bottou2010large,yuan2016convergence,mcmahan2017communication,li2019convergence,li2020federated,karimireddy2020scaffold}. SGD works by iteratively updating the model as
\begin{equation}
x^{t+1}=x^t-\eta\nabla f(x^t;\xi^t),   
\end{equation}
starting from an arbitrary point $x^0$, where $\nabla f(x^t;\xi^t)$ is the stochastic gradient at $x^t$ estimated on $\xi^t$. Here, $\xi^t$ can be a single data, or a minibatch of data uniformly sampled from the whole training data at random.
FedAvg~\cite{mcmahan2017communication} is an extensively employed adaptation of SGD to FL settings.
In FedAvg, the LocalOPT in Algorithm~\ref{alg1} adopts SGD and the server takes the weighted average of local gradients to update the global model.
It has been demonstrated that FedAvg can converge at a sub-linear rate of $\mathcal{O}(\frac{\sigma}{\sqrt{KT}})$ when using a decaying learning rate~\cite{stich2019error,yu2018parallel}.

Despite the statistically optimal convergence rate, SGD suffers from the problem of converging to saddle points and difficulty of adjusting its learning rate. These drawbacks are also inherited by FedAvg. Momentum-based SGD methods~\cite{karimireddy2021learning,liu2020accelerating,huo2020faster,xu2021fedcm,sun2024role} can accelerate model convergence and escape saddle points by incorporating historical gradient information into the current gradient. \cite{yang2022federated} has shown that introducing Nesterov's method into FL leads to the same theoretical convergence rate as FedAvg. However, when the learning rate satisfies mild conditions, momentum-based SGD achieves faster actual convergence speed. Building upon this, a recent study~\cite{reddi2020adaptive} has further integrated typical methods for adaptive learning rate into momentum-based SGD to achieve higher convergence speed. It has been demonstrated that integrating Adam~\cite{hinton2012neural} into FL to optimize the global learning rate $\eta_g$ can achieve the convergence rate of $\mathcal{O}(\frac{\sigma}{\sqrt{KT}})$ in non-i.i.d. scenarios, matching the rate of FedAvg. Variance reduction methods~\cite{liang2019variance, karimireddy2020scaffold, murata2021bias, gao2022feddc} can also be integrated into federated frameworks to reduce the gradient variance term $\sigma$ in FedAvg's convergence rate. Additionally, the Alternating Direction Method of Multipliers (ADMM) algorithm~\cite{zhou2023federated,jeon2021privacy,kant2022federated} can be used to handle constrained federated optimization problems.  Research~\cite{zhou2023federated} has shown that ADMM achieves a convergence rate of $\mathcal{O}(\frac{1}{T})$ when dealing with topology-constrained federated optimization problems in decentralized FL.

\subsection{Second-order optimization}

Second-order optimization methods~\cite{agarwal2017second,sun2019survey} invest additional computational resources to compute the Hessian matrix~\cite{boyd2004convex}, thereby offering a detailed representation of the local optimization landscape. Despite the increased computation and communication overhead, they can deliver a more precise optimization path and often surpass first-order methods with faster convergence speed. The mainstream second-order algorithm in FL is the Newton's method ~\cite{elgabli2022fednew,safaryan2021fednl,dinh2022done,nagaraju2023fonn}, which typically updates the model parameters by combining the Hessian matrix (e.g.,$\nabla^2 f(x^t)$) and the first-order gradient information:
\begin{equation}
x^{t+1}=x^t-\nabla^2 f(x^t)^{-1}\nabla f(x^t).   
\end{equation}
Considering the aggregation process in FL, i.e., $\nabla f(x^t) = \frac{1}{N} \sum_{i=1}^N\nabla f_i(x^t)$, the server updates the global model by 
\begin{equation}
x^{t+1}=x^t-\Big(\frac1N\sum_{i=1}^N\nabla^2f_i(x^t)\Big)^{-1}\Big(\frac1N\sum_{i=1}^N\nabla f_i(x^t)\Big).   
\end{equation}
However, as mentioned, the adaptation of Newton's method to FL suffers from heavy communication and computation of the Hessian matrix. To relieve the burden, one intuitive approach is to use advanced compression methods such as the top-$k$ selection technique \cite{safaryan2021fednl}. 
Furthermore, to avoid computing the inverse of the Hessian matrix, the server can leverage ADMM to approximate Hessian inverse-gradient product $\nabla^2 f(x^t)^{-1}\nabla f(x^t)$ (also called global Newton direction)~\cite{elgabli2022fednew}. 
Some other iterative methods, like Richardson iteration~\cite{dinh2022done}, can also be applied to get local Newton direction, i.e., $\left(\nabla^2f_i(x^t)\right)^{-1}\nabla f_i(x^t)$, which is regarded as a solution of a linear system. It has been proved that FL with these adapted Newton's methods can converge at the quadratic speed as centralized Newton's methods~\cite{elgabli2022fednew,dinh2022done}.

\subsection{Zeroth-order optimization}
The gradient (including the first-order and second-order) information may be inaccessible or computationally prohibitive in some scenarios, e.g., black box models~\cite{guidotti2018survey} and reinforcement learning~\cite{mania2018simple,zhan2023policy}. Zeroth-order (ZO) optimization (also called derivative-free optimization) methods~\cite{conn2009introduction,liu2020primer} works for such scenarios by utilizing the zeroth-order information, i.e., the variation of objective function value along some directions over a mini-batch of data samples. They can naturally be adopted in FL and named as federated ZO~\cite{fang2022communication,haddadpour2019local,qiu2023zeroth}.
Particularly, in federated ZO, the gradients can be approximated by the variation of objective function value as
\begin{equation}\label{eq:zoo}
    \widetilde{\nabla} f_i \left( x_i^t; \{\xi_{i,m}^t\}_{m=1}^{b_1}, \{ v_{i,l}^t \}_{l=1}^{b_2}, \mu \right) = \frac{1}{b_1 b_2} \sum \limits_{m=1}^{b_1} \sum \limits_{l=1}^{b_2} \frac{d v_{i,l}^t}{\mu}(f_i(x_i^t+\mu v_{i,l}^t; \xi_{i,m}^t) - f_i(x_i^t; \xi_{i,m}^t)), 
\end{equation}
where $\{\xi_{i,m}^t\}_{m=1}^{b_1}$ is a set of i.i.d. random samples, $\{ v_{i,l}^t \}_{l=1}^{b_2}$ is a set of i.i.d. random direction vectors (sampling from the $d$-dimensional uniform distribution), and $\mu$ is a positive step size. Then the local model update in LocalOPT of federated ZO algorithms can be written as
\begin{equation}
    x_i^{t+1} = x_i^t - \eta_l  \widetilde{\nabla} f_i \left( x_i^t;\{ \xi_{i,m}^t\}_{m=1}^{b_1}, \{ v_{i,l}^t \}_{l=1}^{b_2}, \mu \right).
\end{equation}

Although federated ZO algorithms adopt a similar framework to FedAvg, the convergence analysis of federated ZO algorithms is slightly different from that of FedAvg. That is because that the gradient estimator of ZO algorithms does not preserve the unbiasedness of stochastic gradients, i.e., $\widetilde{\nabla} f_i \left( x_i^t;\{ \xi_{i,m}^t\}_{m=1}^{b_1}, \{ v_{i,l}^t \}_{l=1}^{b_2}, \mu \right)$ is a biased approximation to the real gradient $\nabla f_i (x_i^t)$~\cite{liu2020primer,fang2022communication}. Therefore, in the analysis of federated ZO algorithms, the gradient estimator is usually decomposed into two components, i.e., the difference between the real gradient and its expected estimator, the divergence between the expected and its ZO-approximated estimator~\cite{chen2024fine}. It has been shown that federated ZO algorithms can achieve sub-linear convergence under the assumptions of non-smooth convex loss functions in both i.i.d. and non-i.i.d. settings~\cite{haddadpour2019local}. Notably, under the assumption of $L$-smooth local objective functions, federated ZO algorithms can also achieve sub-linear convergence in the non-convex setting~\cite{qiu2023zeroth}.

\subsection{Discussion}

In federated optimization, one criterion for selecting appropriate optimization methods is the trade-off between iteration complexity and communication cost~\cite{qian2022basis}. A larger volume of communication data, such as the Hessian matrix, can reduce the total number of iterations, leading to faster model convergence. On the other hand, first-order methods with higher iteration complexity, often incur lower communication costs. Zeroth-order methods are primarily employed in specific FL scenarios where gradients are unavailable, showing no significant advantage in communication or iteration complexity.

\section{Federated Learning with Non-i.i.d. Data}
\label{sec:non-i.i.d.}

\begin{table}[!htbp]
\centering
\caption{Non-i.i.d. classification based on different feature and label distributions~\cite{kairouz2021advances}.}
\label{tab:non-i.i.d.-categories}
\begin{tabularx}{\textwidth}{cc}
\toprule
  \textbf{Non-i.i.d. categories}  &  \textbf{Probability distributions} \\
  \midrule
   Feature distribution skew & $\mathcal{P}_i(y|x)=\mathcal{P}_j(y|x)$, $\mathcal{P}_i(x)\neq \mathcal{P}_j(x)$.  \\
   Label distribution skew & $\mathcal{P}_i(x|y)=\mathcal{P}_j(x|y)$, $\mathcal{P}_i(y)\neq \mathcal{P}_j(y)$. \\
   Varying features under one label & $\mathcal{P}_i(y)=\mathcal{P}_j(y)$, $\mathcal{P}_i(x|y)\neq \mathcal{P}_j(x|y)$.  \\
   Identical features with different labels & $\mathcal{P}_i(x)=\mathcal{P}_j(x)$, $\mathcal{P}_i(y|x)\neq \mathcal{P}_j(y|x)$.   \\
   Data imbalance & {Data amount significantly varies across clients.}  
  \\
\bottomrule
\end{tabularx}
\end{table}

In FL, data is generated naturally and stored locally, giving rise to data heterogeneity, i.e., non-independent and identically distributed (non-i.i.d.) data across diverse clients. Formally, assume any data sample $(x, y)$ of the $i$-th client is drawn from a distribution $\mathcal{P}_{i}(x, y)$, where $x$ and $y$ denote the feature vector and label respectively. $\mathcal{P}_{i}(x, y)$ can be rewritten as
\begin{equation}
    \mathcal{P}_{i}(x, y) = \mathcal{P}_{i}(y|x) \mathcal{P}_{i}(x) = \mathcal{P}_{i}(x|y) \mathcal{P}_{i}(y), \quad i \in [N].
\end{equation}
Then the non-i.i.d. settings in FL can be classified into five categories according to different feature and label distributions~\cite{zhu2021federated, kairouz2021advances}, which are summarized in Table~\ref{tab:non-i.i.d.-categories}.

\subsection{Negative effect of non-i.i.d. data on FL}
In real-world FL tasks, the data heterogeneity comes from the complicated mixture of the above five non-i.i.d. issues. 
Regardless of the category, data heterogeneity inherently leads to the decomposition of the global gradient into multiple biased local gradients over non-i.i.d datasets. The biases in local gradients then compromise the convergence rate and error bound of the global model through the aggregation process~\cite{karimireddy2020scaffold, li2019convergence}. In the following, we will present some theoretical results as well as an intuitive explanation for the performance differences in model convergence between i.i.d. and non-i.i.d. scenarios.

\subsubsection{Theoretical results on FL with non-i.i.d. data}
In the following, we highlight the degraded model convergence by comparing the theoretical convergence rates and error bounds in both i.i.d and non-i.i.d. settings.

\begin{itemize}
\item\textbf{Faster convergence to the first-order stationary point in i.i.d. settings.}  Under the assumption of the smooth objective function, the convergence results are as follows:

\begin{itemize}
    \item \textbf{For strongly convex objectives},  traditional distributed algorithms such as local SGD~\cite{Stich_2018} demonstrate a convergence rate of $\mathcal{O}\left( \frac{1}{T} \right)$ towards the optimal point~\cite{Stich_2018, Khaled_Mishchenko_2020}. 
    \item \textbf{For non-convex objectives}, the model may get stuck in an extreme point or a saddle point, and require more iterations to escape these points~\cite{chen2024escaping}. Therefore, traditional distributed algorithms converge towards the first-order stationary point at a sub-optimal convergence rate of $\mathcal{O} \left( \frac{1}{\sqrt{T}} \right)$~\cite{yu2019parallel}.
\end{itemize}

\item\textbf{Slower convergence to a neighborhood of the first-order stationary point in non-i.i.d. settings.}  
In the theoretical analysis, the non-i.i.d. issue is always characterized by the assumption of $(G, B)$-bounded gradient dissimilarity~\cite{karimireddy2020scaffold}. By further imposing the smoothness assumption on the objective function, we can obtain the following analytical convergence results:

\begin{itemize}
  \item \textbf{For strongly convex objectives}, the error bound of FedAvg is $\mathcal{O} \left(\frac{1}{\mu T} + \frac{G^2}{\mu T} + \frac{B^2}{\mu} \right)$, where $\mu$ is a coefficient to describe the convexity~\cite{karimireddy2020scaffold}. This implies that although FedAvg also achieves a convergence rate of $\mathcal{O}(\frac{1}{T})$, non-i.i.d. data distributions still result in a reduced convergence rate and an increased convergence error due to the existence of terms $\frac{1+G^2}{\mu T}$ and $\frac{B^2}{\mu}$ respectively. 
  \item \textbf{For non-convex objectives}, FedAvg exhibits a sub-linear convergence rate of $\mathcal{O} \left( \frac{1+G^2}{\sqrt{T}} + \frac{B^2}{T}  \right)$~\cite{karimireddy2020scaffold}. 
\end{itemize}  
\end{itemize}

\subsubsection{Intuitive explanations}

\begin{figure}[!htbp]
\centering 
\includegraphics[width=0.95\textwidth]{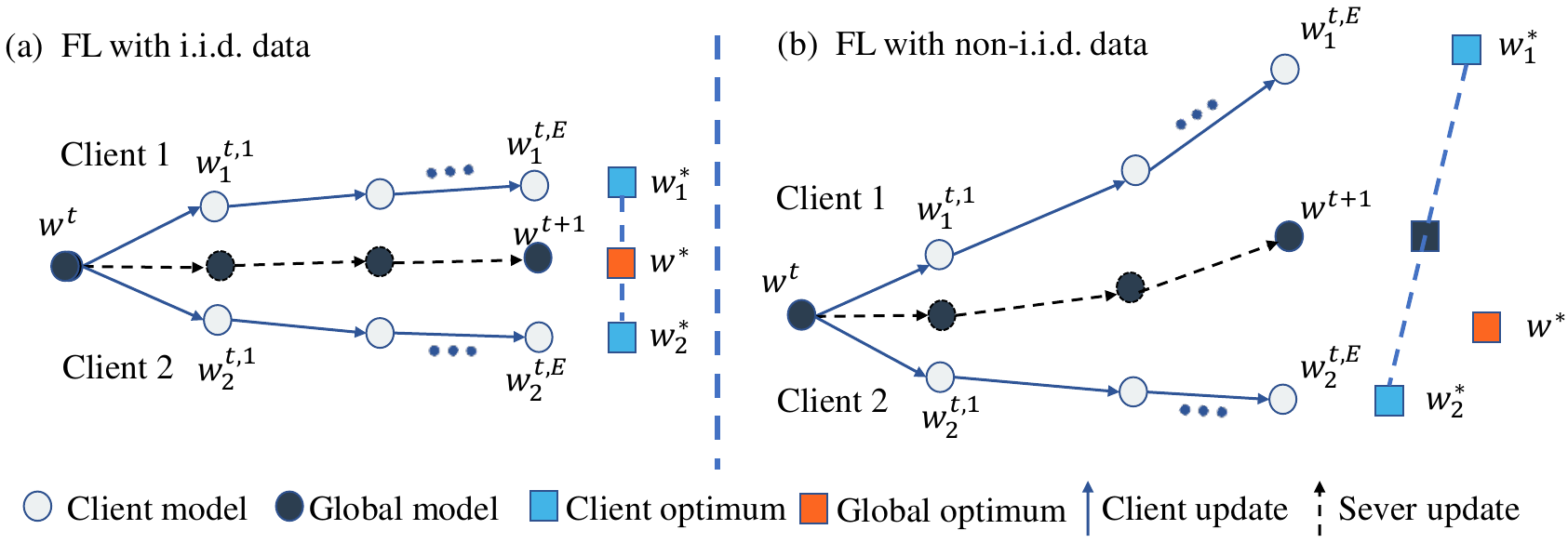} 
\caption{ Illustration of model update trajectories under i.i.d. (a) and non-i.i.d. (b) settings in FL for two clients with $E$ local iterations~\cite{karimireddy2020scaffold}.} 
\label{Fig.4-1} 
\end{figure}

Here we provide an intuitive understanding of the negative impact of non-i.i.d. data on the model convergence by comparing the model updating trajectories in the FL optimization process in both i.i.d. and non-i.i.d. settings.

\begin{itemize}
\item\textbf{Faster convergence to the first-order stationary point in i.i.d. settings.} In ideal i.i.d. settings, as illustrated in \figurename~\ref{Fig.4-1}~(a), the proximity between client optima and global optimum ensures unbiased model updates and consistent convergence under the identical initial models. Consequently, the global model achieves a fast convergence to the first-order stationary point.

\item\textbf{Slower convergence to a neighborhood of the first-order stationary point in non-i.i.d. settings.} In practical non-i.i.d. settings, as illustrated in \figurename~\ref{Fig.4-1}~(b), due to disparities among multiple client optima, FL algorithms such as FedAvg exhibit a decreased convergence rate and increased error bound~\cite{karimireddy2020scaffold}. On the client side, the local model overfits the local data with biased distribution, causing inevitable biases in gradients~\cite{li2019convergence}. This can be characterized as
\begin{equation}
    \mathbb{E} \left[ \nabla f_{i} (x) \right] \neq \nabla f(x), \quad  i \in [N].
\end{equation}
On the server side, the biases cannot be eliminated in the model aggregation and updating process due to the inaccessibility to local data distributions~\cite{li2019convergence}. 
After multiple training rounds, the biases accumulate, which causes the global model to update along a largely biased direction~\cite{wang2020tackling}. Eventually, the non-i.i.d. data significantly degrades the model utility~\cite{zhao2018federated}. 
\end{itemize}

\subsection{Optimization methods for FL with non-i.i.d. data}\label{subsec:non-iid optimization}
The above analysis and explanations show the routine along which the biases in data propagate to the model.
Existing methodologies~\cite{shoham2019overcoming, li2021ditto,t2020personalized,li2020federated} aim to mitigate the non-i.i.d. issue by correcting the biases in different components along the routine. They can be summarized as follows.

\begin{itemize}
\item\textbf{Regularization}~\cite{t2020personalized, li2020federated, li2021ditto,mansour2020three, sun2023dynamic} methods manipulate the objective function to constrain the parameter space, ensuring that the optimal parameters across different clients closely resemble one another. Specifically, regularization methods try to constrain the biases by introducing a regularization term related to the distance between the local and global model parameters. The empirical loss function for the $i$-th client is then represented as
\begin{equation}
    \min_{x} F_{i}(x, x^{t-1}): = f_{i}(x) + \alpha l_{\text{reg}}(x, x^{t-1}),
\end{equation}
where $\alpha$ is the regularization parameter and $l_{\text{reg}}(x, x^{t-1})$ is the regularization term, e.g., the $L_{2}$-regularization term $\| x -x^{t-1}\|^2$~\cite{li2020federated}. 
The regularization term enforces the local model to remain in a limited neighboring region around the global model and reduces the model deviation in the FL optimization process. It has been proved that the loss reduction in each round is lower bounded by $\rho^{t} \| \nabla f(x^{t}) \|^2$ under assumptions of the smooth and non-convex loss function~\cite{li2020federated}, where
\begin{equation}
    \rho= \mathcal{O} \left(\frac{1-\gamma}{\alpha}-\frac{ \gamma (1+\alpha)}{({\alpha}-\theta) \alpha}-\frac{\gamma^2 }{({\alpha}-\theta)^2} \right), 
\end{equation}
$\gamma$ is a parameter positively correlated with the level of data heterogeneity and $\theta$ is a constant. This theoretical result indicates that the convergence accelerates with the increase of $\alpha$ due to the constrained model biases. However, when $\alpha$ exceeds a certain threshold, local models tend to align closely with the received global model and stop learning from local data, thereby slowing down the convergence. By selecting an appropriate $\alpha$, regularization-based algorithms can achieve a convergence rate of $\mathcal{O} (\frac{1}{T})$~\cite{li2021ditto}.

\item\textbf{Model interpolation}~\cite{mansour2020three,yang2023personalized,chen2023fedsoup} combines the biased local model and global model to reduce the distances among different local models, improving the trade-off between model personalization and generalization. 
Specifically, each client trains a composite model $\lambda_{i} x_{i}+\left( 1-\lambda_{i} \right) x$ in each round, where $\lambda_{i}$ is the interpolated weight for the $i$-th client. 
Interpolation facilitates the alignment of each client's model towards the global average model, thereby fostering consistency across different local models throughout the convergence process. Theoretical analysis~\cite{mansour2020three} reveals that the interpolation methods can affect the generalization error bound of each local model $x_i$ through a scaling term $\mathcal{O}\left( \sqrt{\frac{\lambda_i^2}{n_i}+\frac{\left( 1-\lambda_i \right) ^2}{n}} \right)$, where $n_{i}$ and $n$ represent the number of local data and global data, respectively. The generalization error bound scales markedly without interpolation (i.e., $\lambda_i$ equals 0 or 1). Instead, by setting an appropriate $\lambda$, model interpolation may significantly reduce the generalization error bound.

\item\textbf{Adaptive optimizer}~\cite{reddi2020adaptive} 
enhances the exploration capability of the optimizer (e.g., SGD in FedAvg) by adaptively adjusting the learning rates of both local and global models, thereby reducing the biases. This is motivated by the theoretical analysis that FedAvg with a fixed learning rate cannot minimize the convergence error incurred by the non-i.i.d. data~\cite{reddi2020adaptive}. The basic idea is to incorporate advanced adaptive optimizers, e.g., Adagrad~\cite{ward2020adagrad}, Yogi~\cite{zaheer2018adaptive}, and Adam, into the FedAvg framework. Taking Adam as an example, the local and global learning rate is set as $\mathcal{O}(\frac{1}{L\sqrt{T}})$ and $\eta_g = \mathcal{O}(\frac{\sqrt{K}}{\sqrt{v_t}+\frac{G}{L}})$ respectively, and the global updating rule is 
\begin{equation} \label{eq:adaptive-adam}
    x^{t+1} = x^t + \eta_g m^t,
\end{equation}
where $m^t$ denotes the momentum and $v^t$ is a time-varying term that combines the historical and current gradient information. Then the convergence rate of the global model in non-i.i.d. settings can reach $\mathcal{O}(\frac{1}{\sqrt{KT}})$, under the assumptions of non-convex and $L$-smooth objective function and $G$-bounded gradients. This rate matches the standard convergence rate in i.i.d. settings~\cite{wang2018cooperative}. 

\item\textbf{Variance reduction}~\cite{liang2019variance, karimireddy2020scaffold, murata2021bias, gao2022feddc} methods reduce the variance of local gradients among clients, enhancing the consistency among each other. They are motivated by some theoretical results that a larger gradient variance induces decreased convergence rate and increased convergence error~\cite{karimireddy2020scaffold}. Variance reduction is always accomplished by fusing the unbiased global gradient with each local biased gradient.
In particular, in each round of local training, each client combines historical unbiased global gradients and newly calculated local gradients to yield the local update~\cite{murata2021bias, gao2022feddc}. The local updating rule for the $i$-th client is:
\begin{equation}\label{4-6}
   x_{i}^{t+1} = x_{i}^{t}-\eta_l \left( \nabla f_{i}\left( x^t \right) - \nabla f_{i}\left( x^{t-1} \right) +\frac{1}{N}\sum_{i=1}^N{ \nabla f_{i}\left( x^{t-1} \right)} \right) .
\end{equation}
Theoretical results reveal that applying variance reduction methods in non-i.i.d. settings can achieve a similar convergence performance to i.i.d. settings. That is, the convergence rate of $\mathcal{O} \left( \frac{1}{T} \right) $ for strongly convex objective functions and a sub-linear rate of $\mathcal{O}\left( \frac{1}{\sqrt{T}} \right)$ for non-convex objective functions~\cite{karimireddy2020scaffold}.

\item \textbf{Momentum}~\cite{cheng2024momentum, liu2020accelerating, Wang2020SlowMo} utilizes both the historical information and current model to correct the update direction, thereby alleviating the bias induced by non-i.i.d. data. Specifically, this optimization method aggregates the last round gradients and the current update by assigning an appropriate weight. Existing research~\cite{cheng2024momentum, liu2020accelerating} has demonstrated that momentum can alleviate client drift and accelerate the convergence of FL. This is attributed to the fact that the accumulated historical information can enhance the representativeness of data from multiple clients and mitigate the inherent biases stemming from non-i.i.d. data. Specifically, the momentum method can achieve a convergence rate of $\mathcal{O}(\frac{1}{\sqrt{NET}}+\frac{1}{T})$~\cite{Wang2020SlowMo, cheng2024momentum} under the assumptions of $L$-smooth functions.

\item \textbf{Discrepancy-aware aggregation} methods~\cite{ye2023feddisco} allocate aggregation weights that exhibit a negative correlation with local discrepancy levels. These methods are driven by the theoretical analysis of the expected gradient error. It suggests that, to minimize the upper bound of gradient error in FedAvg, aggregation weights should demonstrate a negative correlation with local discrepancy levels while being positively correlated with dataset size. In the implementation, each client first computes its local discrepancy using the local category distribution. Then, the server assigns distinctive aggregation weights based on these discrepancy values.

\end{itemize}

\begin{table}[ht]
\centering
\caption{Summary of optimization methods for FL with non-i.i.d. data. }
\label{tab:noniid-method}
\begin{tabularx}{\textwidth}{Y{2.8cm}Y{2.5cm}Y{3.9cm}Y{4.1cm}}
\toprule
\textbf{Optimized components}        & \textbf{Methods}                &  \textbf{Advantages }         &   \textbf{Disadvantages}     
\\
\midrule
Objective  function
    & {\centering Regularization}       & {Easy deployment}        &    {Difficulty in selecting appropriate  $\alpha$} 
\\
\midrule
\multirow{4}{2.5cm}{\centering Model }  & {\centering Model interpolation } & {Balance between generalization and personalization}                 &  {High computation complexity and memory consumption}  
\\
\cline{2-4}
 &   Discrepancy-aware aggregation   &   
 Comprehensive theoretical foundation    &   Difficulty in selecting weight-related hyper-parameters    \\
\midrule
Optimizer 
    & {\centering Adaptive optimizers}       & {Enhanced model exploration capability}             &     {High computation complexity and memory consumption}   
\\
\midrule
\multirow{4}{2.5cm}{\centering Gradients}   &  {\centering Momentum}        &  {Enhanced model convergence}             &    { High memory consumption }  
\\
\cline{2-4}
 & {\centering Variance reduction}        & {Tight analysis in non-i.i.d. settings}             &    { High communication cost and memory consumption }   
\\
\bottomrule
\end{tabularx}
\end{table}

\subsection{Discussion}

As summarized in Table~\ref{tab:noniid-method}, the regularization methods are convenient to deploy since they can be achieved by simply manipulating the objective function. However, it is challenging to select an appropriate regularization parameter $\alpha$ in prior. Model interpolation methods can strike a balance between generalization and personalization, but require additional computational resources to learn the interpolation parameter $\lambda$ and extra memory to store the global model for each client. 
Discrepancy-aware aggregation stems from a rigorous theoretical analysis of optimizing error bounds and has a tight theoretical analysis. However, the model performance relies on the properly selected metric for local discrepancy levels and hyper-parameters for discrepancy-aware aggregation weights.
Adaptive optimizers enhance model exploration but introduce complexity, risking bias-variance dilemmas and demanding more computational and memory resources. 
Momentum methods improve the convergence via utilizing the historical information, which however requires additional memory consumption for both clients and the server.
Variance reduction methods offer rigorous theoretical guarantees in non-i.i.d. scenarios. However, they necessitate additional communication costs and memory consumption for transmitting and storing historical gradients.

\section{Federated Learning with Differential Privacy}\label{sec:dp}

Differential Privacy (DP) provides a mathematical framework to formulate and control privacy loss in FL~\cite{zhang2022understanding, girgis2021shuffled}. Despite no raw data exchange, the communicated models in FL may still disclose sensitive information about training data. By incorporating the DP constraint~\cite{ullman2018tight,kuru2022differentially,chen2024locally}, the FL optimization can be regularized to converge to a model with limited information disclosure.

Formally, let $\mathcal{M}$ be a differentially private FL (DP-FL) algorithm that takes the distributed dataset $\mathcal{D} = \{\mathcal{D}_1, \ldots, \mathcal{D}_N\}$ as input and outputs the well-trained model $x$.  Then $\mathcal{M}$ satisfies $(\varepsilon, \delta)$-differential privacy if it holds that 
\begin{equation}\label{equ:dp}
    Pr[\mathcal{M}(\mathcal{D}) = x]\le e^{\varepsilon}Pr[\mathcal{M}(\mathcal{D}^{\prime})= x] + \delta,
\end{equation}
where $\mathcal{D}^{\prime}$ is a virtual dataset which differs in $\mathcal{D}$ with only one record. According to the different granularity of $\mathcal{D}^{\prime}$, the privacy guarantee can be classified into \textit{sample-level} (i.e., $\mathcal{D}^{\prime}$ differs from $\mathcal{D}$ in a single sample~\cite{yuan2021beyond}) and \textit{client-level} (i.e., $\mathcal{D}^{\prime}$ differs from $\mathcal{D}$ in a client's dataset~\cite{yuan2021beyond}). 

Besides the granularity of privacy protection, it is also necessary to consider different adversary/threat models. According to whether the server is trusted or not, there are two main DP models, namely centralized DP (CDP) and local DP (LDP)~\cite{wang2017locally,li2023robustness,wu2022federated}. In CDP, the server is assumed trustworthy, while the threat comes from external malicious analysts or clients. These adversaries may have access to global models from the server to infer the sample or client-specific information. To address this, the server introduces noise into the global model to safeguard privacy in FL with CDP. In contrast, in LDP, the server is assumed honest-but-curious, meaning it follows the protocol but may attempt to infer the information of training samples via the received intermediate results from clients. In this case, FL clients have to locally perturb their gradients or models to defend against this threat in FL with LDP~\cite{wang2023ppefl,batool2024secure,wang2020federated,wu2022federated}. Note that, with the same privacy parameter $\epsilon$, LDP provides stronger privacy protection than CDP, but causes larger utility loss. To improve the privacy-utility tradeoff, distributed DP (DDP)~\cite{cheu2019distributed} integrated with secure shuffling~\cite{girgis2021shuffled} or secure aggregation~\cite{chen2022fundamental, agarwal2018cpsgd} has also been proposed, in which the local model updates are perturbed with a smaller noise and then shuffled or securely aggregated to achieve a sufficient CDP noise.

\begin{figure}[htbp]
\centering 
\includegraphics[width=0.75\textwidth]{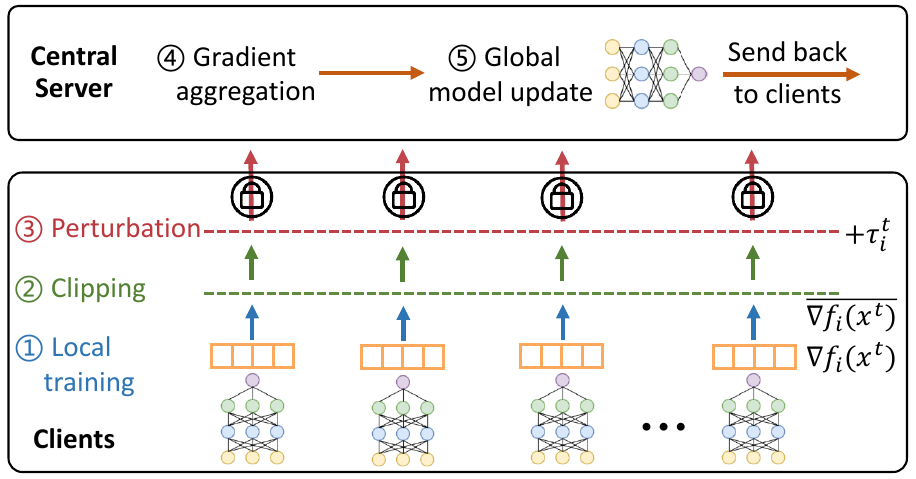}
\caption{Typical workflow of FL with DP.}
\label{fig:DPframework}
\end{figure}
We use the sample-level DP as the default setting to illustrate the general workflow of FL with DP. As shown in Fig.~\ref{fig:DPframework},
each client first conducts local training and then clips the local gradients $\nabla f_{i}(x^{t})$ with a positive constant $c$:
\begin{equation}\label{5:5}
\overline{\nabla f_{i}(x^{t})}=\nabla f_{i}(x^{t}) \cdot \min \bigg\{1, \dfrac{c}{ \| \nabla f_{i}(x^{t}) \|_2 }\bigg\}.
\end{equation}
\textit{Gradients clipping}~\cite{abadi2016deep} is a common method to limit the gradients' \textit{sensitivity}, i.e., the change of the gradient norm due to adding or deleting an individual sample. Afterward, a certain level of DP can be achieved by injecting carefully calibrated noises (e.g., Gaussian~\cite{abadi2016deep} or Laplacian~\cite{wu2020value} noises) into the clipped gradients. Taking the most commonly used Gaussian mechanism as an example, the scale of Gaussian noise is specified as $\sigma^2=\frac{2c^2 \log(1.25/\delta)}{\epsilon^2}$ to achieve $(\epsilon, \delta)-$DP~\cite{abadi2016deep}.  Finally, the perturbed gradient $ \overline{\nabla f_{i}(x^{t})}+ \tau_i^t $ is sent to the central server for updating the global model~\cite{yuan2021beyond}.

\subsection{Negative effect of DP on FL optimization}\label{sec:dp-effect}

As discussed, realizing DP in FL involves the operations of gradient clipping and noise addition to the vanilla FL algorithms. While guaranteeing DP, these operations inevitably lead to a reduced convergence rate and increased error bound, reflecting the privacy-utility trade-off~\cite{hu2023federated, girgis2021shuffled}. Specifically, their major impacts on the model convergence can be summarized as follows:

\begin{itemize}

\item \textbf{Decreased convergence rate.} Both gradient clipping and noise addition introduce addition error, thus impacting the convergence of FL algorithms~\cite{zhang2022understanding, hu2023federated, girgis2021shuffled}. As analyzed in~\cite{girgis2021shuffled}, the convergence rate of DP-enhanced FL can be represented as
\begin{equation}\label{eq:dp-effect2}
    \mathcal{O}\left(\frac{\log(T)\max\{d^{\frac12-\frac1p},1\}}{\sqrt{T}}\sqrt{\frac{cd}{q}}\left(\frac{e^{\epsilon}+1}{e^{\epsilon}-1}\right)\right),
\end{equation}
where $p$ represents the norm exponent adopted to clip the gradient, $q$ is the data sampling ratio in a mini-batch, and $d$ is the dimension of model parameters. Specifically, it is This theoretical result indicates that a larger sampling ratio $q$ and privacy budget $\epsilon$ can speed up the convergence, while the larger gradient dimension $d$ and clipping bound $c$ can slow down the convergence.

\item \textbf{Increased error bound.} Besides the convergence rate, DP also harms the final model utility. By achieving DP in FedAvg, an error bound of the trained model~\cite{zhang2022understanding} can be derived as follows with a small clipping bound $c$.
\begin{equation}\label{5:4}
    \begin{aligned}
    \frac{1}{T}\sum_{t=1}^{T}\mathbb{E}\left[{\alpha}^{t}\left\|\nabla f\left(x^{t}\right)\right\|^{2}\right] \leq \mathcal{O}\left( \frac{1}{\sqrt{KET}} \right)+\mathcal{O}(c^2E)+\mathcal{O}(\frac{d\sigma^2}{EK}), \\
    \end{aligned}
\end{equation}
where $\alpha$ is a constant, $K$ is the number of participant clients, and $\sigma^2$ and $d$ represent the noise scale and the dimension respectively. The term $\mathcal{O}(c^2E)$ in the right side of Eq.~\eqref{5:4} captures the impact of clipping, wherein a larger clipping bound $c$ indicates a higher error bound. Similarly, the last term $\mathcal{O}(\frac{d\sigma^2}{EP})$ demonstrates that the error bound increases linearly with the dimension $d$ and noise scale $\sigma^2$.
\end{itemize}

\subsection{Optimization methods of FL with DP}\label{sec:dp-methods}

Section~\ref{sec:dp-effect} reveals that an additional DP guarantee compromises the model utility in terms of both error bound and convergence rate. Recent studies~\cite{wei2020federated, cheng2022differentially} propose several strategies to improve the model utility in FL without sacrificing the DP guarantee, including the clipping bound calibration to reduce sensitivity~\cite{li2023multi}, sampling for amplifying the privacy preservation~\cite{girgis2021shuffled}, and gradient sparsification to enhance the siginal-noise-ratio~\cite{cheng2022differentially}.

\begin{itemize}
\item\textbf{Sensitivity reduction by fine-grained clipping.}
As analyzed in Section~\ref{sec:dp-effect}, a small clipping bound can effectively improve the model utility through limiting the added noises. However, an undersized clipping bound $c$ can lead to severe biases in the aggregated results~\cite{chen2020understanding,li2023multi}, also diminishing the model utility. Therefore, the clipping bound should be calibrated carefully.
To this end, recent study~\cite{li2023multi} empirically observes the reduction of gradient norms with increasing $T$ and accordingly proposes an adaptive norm-aware clipping algorithm. The fine-grained clipping algorithm effectively reduces the aggregation bias and DP noises,  achieving a convergence rate of $\mathcal{O} (\frac{1}{T})$ under the assumption of smooth objective function~\cite{li2023multi}.

\item\textbf{Privacy amplification by sampling.}
As a common method, sampling~\cite{shi2023make,girgis2021shuffled}, e.g., the client sampling and data sampling in FedAvg, can amplify the privacy protection capability of a randomized FL algorithm by introducing additional randomness. 
Given $\epsilon_{0}$ as the privacy budget to protect gradients in each round, the total $T$ rounds of training can be proved to satisfy $(\mathcal{O}(\epsilon_{0} \sqrt{qT\log(qT)}), \delta)$-DP~\cite{girgis2021shuffled}, where $q$ is the sampling ratio. This theoretical result indicates that by setting a smaller sampling ratio, the integration of the sampling method can significantly reduce the DP noises required to perturb the gradients~\cite{wei2020federated}, thereby improving the model utility. However, this is conflicted with the analysis that a larger sampling ratio speeds up the convergence, as discussed in Section~\ref{sec:dp-effect}. 
Therefore, we conclude that there exists a trade-off between the model convergence and the sampling ratios~\cite{girgis2021shuffled}. Thus, through appropriate calibration of the sampling ratios, substantial enhancements in model utility can be achieved.

\item\textbf{Dimension reduction by sparsifying the gradients.}
Eq.~\eqref{eq:dp-effect2} demonstrates that a high dimension $d$ of gradients can degrade the model convergence ~\cite{dai2022dispfl}. 
A direct solution involves randomly perturbing $k$ out of $d$ dimensions in the model gradients with Gaussian noise while setting the remaining dimensions to zero~\cite{hu2021federated}, based on which, the dimension-related terms in the error bound (Eq~\eqref{5:4}) can then be scaled by $\frac{k}{d}$, thus decreasing the error bound. However, such a random sparsification algorithm introduces significant biases into gradients, thus also degrading the model utility. To this end, recent work~\cite{cheng2022differentially} first discovers the effect of zeroing out different dimensions of gradients on the model utility, and accordingly proposes to identify and zero out $d-k$ dimensions with the least impact.

\end{itemize}

\subsection{Discussion}
    The core challenge of FL optimization with DP is to strike a satisfactory balance between privacy protection and model performance (e.g., model errors and convergence rates). Under a certain level of privacy protection, reducing the intensity of DP noises is a core idea to achieve better model performance. Generally, the noise intensity is proportional to both the model sensitivity and the size of communicated data in the whole training process~\cite{li2022differentially,qi2023differentially}. Given this fact, optimization methods described in Section~\ref{sec:dp-methods} utilize gradient clipping to reduce model sensitivity, and employ gradient sparsification or sampling to reduce the size of communicated data, thereby decreasing the added noise.
    Additionally, sampling, which introduces extra randomness, further enhances the level of privacy protection. Besides, these methods are complementary and can be combined to achieve a better trade-off between privacy protection and model performance.

\section{Federated Learning with Decentralized Topologies}
\label{sec:dfl}

\begin{figure}[!t]
  \centering
  \subfigure[Fully connected topology.]{\includegraphics[width=0.31\textwidth,trim=120 125 680 303,clip]{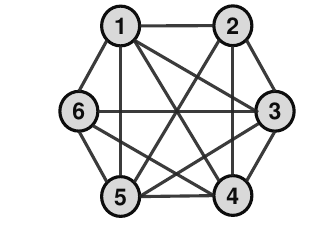}}
  \hfill
  \subfigure[Partially connected topology.]{\includegraphics[width=0.31\textwidth,trim=320 125 480 303,clip]{pic/dfl/DFL-topology.pdf}}
  \hfill
  \subfigure[Time-varying connected topology.]{\includegraphics[width=0.36\textwidth,trim=510 127 265 290,clip]{pic/dfl/DFL-topology.pdf}}
  \caption{Illustration of network topology for fully connected topology (a), partially connected topology (b), time-varying connected topology (c). }\label{sec:dfl:arc}
\end{figure}

Conventional FL, which is also named server-client FL~\cite{zhou2024trustbcfl}, enables a central server to coordinate the learning task by communicating with multiple distributed clients. However, server-client FL encounters several problems, such as single-point failure~\cite{qu2021decentralized}, communication bottleneck~\cite{hashemi2021benefits}, and potential malicious server~\cite{gao2023decentralized}, etc. As an effective alternative, Decentralized FL (DFL)~\cite{9850408,9524471,10251949} can mitigate these problems.

In DFL, each client can only communicate with its neighboring nodes. Formally, the network topology of DFL can be modeled as an adjacent matrix $\mathbf{A} \in \mathbb{R}^{N\times N}$, where $\mathbf{A}_{i,j} = 1/0$ indicates the presence/absence of the communication link between nodes $i$ and $j$. As shown in \figurename~\ref{sec:dfl:arc}, the network topology can fall into three categories according to different links among nodes, including fully connected topology~\cite{ye2022decentralized}, partially connected topology~\cite{10251949} and time-varying connected topology~\cite{warnat2021swarm,doi:10.1137/17M1151973}.
 
Under fully connected topology, each pair of nodes maintains direct link~\cite{ye2022decentralized}, and the adjacent matrix $\mathbf{A} = \mathbf{E}$, where $\mathbf{E}$ is the all-ones matrix. Under partially connected topology, each node only maintains direct links to a subset of nodes~\cite{10251949} and the adjacent matrix is a time-invariant symmetric $0/1$ matrix. Classical partially connected topologies include ring-structured~\cite{wang2021efficient} and clique topologies~\cite{bellet2022d, vogels2022beyond}, etc. Under time-varying connected topology, links among nodes are randomly or selectively determined based on specific factors~\cite{warnat2021swarm} such as resource availability. Here, the adjacent matrix is a time-varying symmetric $0/1$ matrix.

In DFL with the above topologies, due to the lack of a central server and limited communication range, it becomes challenging to obtain global gradient information for each client in each round of aggregation. Several DFL methods have been proposed to mitigate this issue for different underlying topologies. Some typical ones are summarized as follows.

\begin{itemize}
    \item \textbf{Decentralized FedAvg}~\cite{gao2023decentralized} is an adaptation of FedAvg in server-client FL. It enables each node, e.g., the $i$-th node, to aggregate neighboring gradients to update its own model $x_i$. Through multiple local aggregations, $x_i$ can gradually fuse the information from the nodes that is not in $\mathcal{N}_i$.  Such information propagation in the decentralized network can facilitate the model consistency and convergence.
    \item \textbf{Cyclic learning}, originally proposed in decentralized and consensus optimizations \cite{mao2020walkman, mao2018walk}, has been also adapted into DFL~\cite{CRIADO2022263, sheller2020federated}. It allows different clients to train a single model sequentially and cyclically in the FL system with ring-structured topology. Specifically, in each round, each client trains the model received from the previous client and then passes the trained model to the next client. Such a process proceeds for multiple rounds. However, cyclic learning may suffer from poor model utility due to \textit{catastrophic forgetting}~\cite{wang2024comprehensive,wang2023federated,dong2023no}.
    \item \textbf{Swarm learning}~\cite{warnat2021swarm, saldanha2022swarm} dynamically elects the leader among nodes to aggregate local models based on the consensus mechanism of blockchain. It combines the strengths of server-client FL and DFL, eliminating the need for a trusted third party while guaranteeing consistency among local models. It can achieve the same model utility as the server-client FL with additional computation and communication costs in the elected leader in each round.
\end{itemize}

\subsection{Impact of different topologies on FL}\label{subsec:dfl-effect}
Decentralized FL is essentially an extension of the well-established field of decentralized optimization~\cite{nedic2009distributed, lopes2008diffusion}. In traditional decentralized optimization, the influence of topologies has been revealed explicitly~\cite{koloskova2020unified}. And it is apparent that the connectivity of the graph topology impacts the model performance in DFL~\cite{9850408, zhang2022net} as well. However, it encounters new challenges in analyzing the impact of decentralized topologies on DFL due to the distinct characteristics of FL scenarios, such as constrained communication, heterogeneous data, privacy concerns, etc. Specifically, the lower node connectivity coupled with non-i.i.d. dataset, even DP noise,  will significantly hinder the information propagation, and degrade the model convergence and generalization performance~\cite{yuan2016convergence, doi:10.1137/21M1465081,zhang2022net}.

\begin{itemize}
\item \textbf{Decreased convergence rate.} It has been proved that static topology can achieve a convergence rate of $\mathcal{O}(\frac{1}{\sqrt{T}}+\frac{1}{(1-\lambda)^2T^{3/2}})$~\cite{9850408}, where $\lambda \in (0,1)$ is the second largest absolute value of the eigenvalues of the mixing matrix (i.e., maximum-degree matrix and metropolis-hastings matrix~\cite{boyd2004fastest}) associated with network topology. A smaller $\lambda$ corresponds the higher graph connectivity of the topology.  The convergence rate reveals that a smaller $\lambda$ can result in a faster convergence~\cite{yuan2016convergence,9850408,doi:10.1137/21M1465081,liang2023gradienttrackinghighdimensional}, meaning that a higher graph connectivity can facilitate the model convergence. A randomly connected topology probably corresponds to a larger $\lambda$, thereby decreasing the model convergence rate. Furthermore, as a common issue in FL, the data heterogeneity is also proven to be an important factor to degrade the model convergence in DFL~\cite{9850408, koloskova2020unified,pmlr-v139-esfandiari21a}. 

\item\textbf{Decreased generalization ability.} 
It has been proved that the \textit{generalization error} of DFL, i.e., the difference between the true risk of model on the real data distribution and the empirical risk on the training data distribution, is a monotone-increasing function with respect to $\lambda$. This implies that the
generalization ability of DFL can be degraded by the low connectivity of the underlying topology~\cite{yuan2016convergence, doi:10.1137/21M1465081}. 

\end{itemize}

\subsection{Optimization methods for DFL}

Extensive optimization methods for improving the model convergence and generalization ability in DFL have been proposed~\cite{9850408, zhang2022net}. We review these methods from the aspects of topology-aware optimization~\cite{doi:10.1137/21M1465081}, and accelerated optimization~\cite{pmlr-v139-kovalev21a,doi:10.1137/17M1151973}.

\begin{itemize}
  \item \textbf{Topology-aware optimization} methods~\cite{doi:10.1137/21M1465081} attempt to design graph topology with higher connectivity to facilitate model convergence and enhance model generalization ability. However, a highly connected graph inevitably incurs a large overhead on communications~\cite{ying2021exponential, song2022communication}. To speed up model convergence with limited communication resources, existing work proposes a sparse network topology design method based on $d$-regular expander graphs to optimize the trade-off between connectivity and communication, where $d$ is a pre-specified threshold of graph degree. This method gradually densifies the network topology based on a set of virtual coordinates and recursive queries until the degree of each node reaches $d$.

  Considering the topology of DFL, multiple gossip steps (MGS), which means more frequent communication, can result in improved consensus among the participating clients. In view of this, a balance between the communication cost and generalization ability can be ensured~\cite{pmlr-v202-shi23d}.
  \item \textbf{Accelerated optimization} methods~\cite{zhang2022net,pmlr-v139-kovalev21a,doi:10.1137/17M1151973} adapt advanced optimization techniques to facilitate model convergence in DFL settings. By approximating the global unbiased gradient using neighboring local gradients, the variance reduction method corrects the local gradient for each client~\cite{zhang2022net, zhang2023variance,xin2020variance}. This reduces model inconsistency caused by data heterogeneity and achieves a convergence rate of $\mathcal{O}(\frac{1}{\sqrt{KT}})$. Combining the Nesterov gradient descent method with gradient compression, a contractive compression operator is derived for the time-varying decentralized topology~\cite{pmlr-v139-kovalev21a}. Theoretical analysis demonstrates the accelerated model convergence. Besides, by modeling the DFL task as a constrained optimization problem:
    \begin{equation}\label{eq:DFL}
    \begin{aligned}
        & \min_{x_{i}, \ i\in[N]}  \ \ \frac{1}{N}\sum_{i=1}^{N}f_i(x_i),   \\
        & s.t. \ \ x_i=x_j,  \ \ \forall (i,j)\in  \mathcal{E} ,
    \end{aligned}
    \end{equation}
    where $\mathcal{E}$ represents the set of node pairs that maintain links, ADMM can be applied to decouple and solve the DFL optimization problem, effectively enhancing model consistency among nodes~\cite{doi:10.1137/17M1151973, li2022robust}.  Theoretical results demonstrate that the iteration sequence generated by the ADMM algorithm converges to $\epsilon$-suboptimality with a manageable iteration complexity $\mathcal{O}(\frac{1}{\epsilon})$ under time-varying (un)directed network topologies.

    By utilizing sharpness aware minimization (SAM), i.e., introducing a small perturbation to the models, both generalization and robustness are enhanced~\cite{pmlr-v202-shi23d}. Meanwhile, the improved convergence rate $\mathcal{O}(\frac{1}{\sqrt{KT}}+\frac{1}{T}+\frac{1}{K^{1/2}T^{3/2}(1-\lambda)^2})$ can be achieved in the non-convex settings.
\end{itemize}

\begin{table}
\centering
\caption{Summary of optimization methods in DFL.}
\label{table:dfl-method}
\begin{tabularx}{\textwidth}{Y{2.4cm}Y{4.7cm}Y{3.8cm}Y{2.4cm}}
\toprule
{\textbf{Methods}}  & {\textbf{Advantages}} &  {\textbf{Disadvantages}}   &   {\textbf{Network topologies}} 
\\
\midrule
{Topology-aware optimization} &  {Enhanced model convergence and generalization} & {Constraints of connectivity among clients} & {Partially connected topology} 
\\
\midrule
{Accelerated optimization} & {Enhanced model convergence} &  {High computation cost} & {All topologies}  
\\
\bottomrule
\end{tabularx}
\end{table}

\subsection{Discussion}

Table~\ref{table:dfl-method} summarizes the advantages and disadvantages of the two optimization methods. Topology-aware optimization methods improve the model convergence and generalization ability by optimizing the underlying topology directly. However, it is only applicable to settings where the topology is allowed to change. In practice, the generated topology potentially imposes higher communication costs on neighboring nodes located at substantial geographic distances. Accelerated optimization methods can enhance the model convergence by manipulating the gradients in each round of training. It is applicable to all types of topologies but suffers from relatively higher computation costs to correct the gradients in each round.

\section{Online Federated Learning}
\label{sec:online}

The above discussions of FL optimization assume that each client possesses a fixed and static dataset, which is also referred to as batch-based FL. 
However, in numerous real-world scenarios, the training data of clients are often generated in a streaming mode~\cite{banerjee2007topic}. Consequently, online FL (OFL)~\cite{hong2021communication,chen2020asynchronous,m2022personalized,kwon2023tighter} has been introduced by combining online learning (OL)~\cite{hazan2016introduction, hoi2021online} and FL. OFL performs online optimization in FL over the time-evolving data, and makes a prompt label prediction or decision upon receiving incoming data.

OFL aims to learn a sequence of global models from distributed streaming data at local devices. 
At timestamp $t$, each client indexed by $i\in [N]$ receives a new data $(u_i^t,v_i^t)$, where $u_i^t$ is the feature and $v_i^t$ is the label, and the latest global model $x^t$ from the server. The global model is used to predict the label of newly incoming data. Thus, the $i$-th client has a local loss $\mathcal{L}(f(u_i^t;x^t),v_i^t)$, where $\mathcal{L}(\cdot,\cdot)$ is a loss function that measures the error between true and predicted labels. Leveraging the local loss, the $i$-th client optimizes its local model $x_i^{t+1}$ and sends it to the server. After receiving all local models, the server updates the global model $x^{t+1}$ by averaging the local models $\{x_i^{t+1}\}_{i=1}^N$. OFL aims to seek a sequence of global models $x^1,x^2,\ldots,x^t$ that minimizes the cumulative regret (i.e., the difference between cumulative loss of the algorithm and that of the static optimal function) over $T$ timestamps:
\begin{equation}
    \text{Regret}_T = \sum_{t=1}^T \sum_{i=1}^N \mathcal{L}(f(u_i^t;x_i^t),v_i^t)-\min_{x\in \mathbb{R}^d}\sum_{t=1}^T \sum_{i=1}^N \mathcal{L}(f(u_i^t;x),v_i^t).
\end{equation}

\subsection{Effect of online paradigm on federated optimization} \label{7-2}
The online paradigm introduces novel optimization problems and requires new optimization methods for handling streaming data. Here, we discuss the effects of the online paradigm on federated optimization from the following two aspects. 

\begin{itemize}

\item \textbf{Exacerbated issue of obsolete solutions.} 
In online settings, the model is required to be updated exclusively with newly acquired data and continuously give solutions for dynamic environments~\cite{hong2021communication, kwon2023tighter}.
However, general FL methods may be unable to update the model continuously according to new data. This produces obsolete solutions, as the data distribution may have changed, leading to poor model performance (e.g., regret). Furthermore, the constraints of computing and communication in FL aggravate the difficulty of continuously updating the model, which causes a more serious issue of obsolete solutions.

\item \textbf{Severe catastrophic forgetting.}
In the scenario of continuously learning a sequence of tasks based on the online learning paradigm, a model may exhibit degraded performance on old tasks if it only learns new tasks when acquiring new data. This is known as the \textit{catastrophic forgetting} issue~\cite{wang2024comprehensive,wang2023federated,dong2023no}. More seriously, different FL clients may have different task sequences, which may lead to the mutual interference of task knowledge between different clients~\cite{wang2023federated,dong2022federated,dong2023no}. This further aggravates the catastrophic forgetting of clients.

\end{itemize}

\subsection{Optimization methods for OFL}
To alleviate the negative effects caused by the online paradigm, many optimization methods for OFL have been proposed~\cite{chen2020asynchronous,hong2021communication,m2022personalized,kwon2023tighter}. We review these methods from the aspects of addressing obsolete solutions and mitigating catastrophic forgetting issues.
\begin{itemize}
\item
\textbf{Kernel-based methods for addressing obsolete solutions}.
Current OFL methods~\cite{hong2021communication,m2022personalized,kwon2023tighter} mainly rely on online gradient descent method (OGD)~\cite{hazan2016introduction}, an adaptation of SGD for online optimization, to update local models.
For better solving non-linear optimization problems (e.g., deep learning tasks) and addressing the issue of obsolete solutions in OFL, some works~\cite{hong2021communication,m2022personalized,gogineniiot2022communication} integrate OGD, FedAvg and multi-kernel learning (MKL). MKL is an advanced method in OL and has exhibited superior performance~\cite{dekel2008forgetron, hoi2013online}. The kernel that maps the original input space to a higher-dimensional feature space can significantly improve the generalization capabilities of the model, thereby addressing the issue of obsolete solutions. 
It is analyzed that the existing OFL methods with multiple kernels can achieve an optimal sub-linear regret $\mathcal{O}(\sqrt{T})$ by setting the learning rate as $\mathcal{O}(\frac{1}{\sqrt{T}})$.

\item 
\textbf{Mitigating catastrophic forgetting at the data and model levels.}
Existing methods mitigate the catastrophic forgetting issue from both data and model perspectives. Data-based methods aim to leverage historic training samples or generated samples with the similar distribution for training models with new classes. One way is to simply store and replay old samples at the client~\cite{yoon2021federated}. The other is to use generative models to simulate training data with the similar distribution to the historic data~\cite{babakniya2024data,zhang2023target}. 
Model-based methods try to balance the model stability on old tasks and generalization on new tasks~\cite{wang2024comprehensive}. For exmaple, weighted averaging of new and old models~\cite{yoon2021federated} and decreasing learning rates~\cite{yuan2023peer} are beneficial in avoiding catastrophic forgetting for local model updates. The specific mitigating methods can be further divided into the regularization-based and knowledge distillation-based ones. The former usually adds penalty terms to the loss function~\cite{yoon2021federated} while the latter applies knowledge distillation to old training samples~\cite{ma2022continual,babakniya2024data,zhang2023target}. 
Regarding the knowledge distillation-based methods, prototype-based learning approaches~\cite{dong2022federated,dong2023no,shenaj2023asynchronous} are often utilized to collect prototype data of each class and monitor global model performance, which helps to choose the best old global model for knowledge distillation to mitigate the impact of inter-client interference. Also, a recent method of parameter decomposition ~\cite{yoon2021federated} separates the network parameters into the global and task-specific parameters, enabling clients to selectively learn from each other. 

\end{itemize}

\subsection{Discussion}
Currently, research on OFL is still in its early stages. The basic framework integrates FedAvg and OGD. Existing works in OFL address obsolete solutions using OL methods like kernel-based approaches~\cite{hong2021communication,m2022personalized,gogineniiot2022communication}. They also tackle the problem of catastrophic forgetting by adopting centralized continual learning methods such as regularization-based approaches ~\cite{yoon2021federated,ma2022continual,babakniya2024data,zhang2023target}. However, OFL research has yet to explore the important aspects of heterogeneity and privacy issues in FL from both challenges and optimization methods perspectives.

\section{Other important works}
\label{sec:other-works}
In addition to FL optimization studies discussed in the above sections, there are other important research topics, such as sparsification methods~\cite{almanifi2023communication,cho2020client} and gradient aggregation rules~\cite{miao2023robust,blanchard2017machine}. Sparsification methods aim to 
optimize the communication complexity of FL systems while maintaining the model performance. Gradient aggregation rules aim to evaluate and aggregate gradients from different local clients to generate a high-quality (e.g., Byzantine robust) global model update.

\subsection{Sparsification}
Reducing communication complexity is another core challenge in FL optimization~\cite{almanifi2023communication,khan2021federated}, especially in the context of massive participants and complex models. Sparsification~\cite{khan2021federated,zhao2023towards,jiang2022towards}, as a class of communication-efficient methods, can cut down the communication costs by reducing the number of exchanged parameters, which can also improve the generalization ability of the model~\cite{khan2021federated}. However, these methods have a significant impact on the convergence rate and errors of the model. 
Thus, the essential problem is to optimize the model convergence while satisfying the communication constraint via calibrating the sparsification parameter.
Next, we will review the advanced studies from three perspectives, including client sparsification, temporal sparsification and gradient sparsification.

\subsubsection{Client sparsification}
Exchanging model updates with abundant participating clients contributes to the communication bottleneck during an FL training round. Client sparsification (also called client sampling)~\cite{mcmahan2017communication,luo2022tackling,fraboni2021clustered}, i.e., a random selection of a subset of clients, is a viable solution, but that randomness may result in a lot of missed potential. In most FL implementations, the clients vary in design and capability, a diversity that extends to the quality of communication mediums. 
Choosing the clients that meet the most favorable communication conditions in each round should help achieve a higher convergence rate~\cite{almanifi2023communication}. 
It is analyzed in~\cite{cho2020client} that a larger selection skew results in faster convergence at the rate $\mathcal{O}(\frac{1}{T\rho})$, where $\rho$ represents the selection skew towards clients with higher local losses, which reveals that biasing client selection towards clients with higher local loss achieves faster convergence.

\subsubsection{Temporal sparsification}
To leverage all the available data samples on the data owners, standard FL methods generally let the clients synchronize their models through the server in each training iteration. However, this implies many rounds of communication between the clients and the server which results in communication contention over the network. Instead, some works~\cite{reisizadeh2020fedpaq,kwon2023tighter} propose that participating clients conduct several local updates and synchronize through the server periodically. Specifically, once clients pull an updated model from the server, they update the model locally by running $\tau$ iterations of the SGD method and then send proper information to the server for updating the aggregate model. Under a strongly convex setting, for a total number of iterations $T = K \tau$, where $K$ is a positive integer, the convergence rate is $ \mathcal{O}(\frac{\tau}{T})$ $+ \mathcal{O}({\frac{\tau^2}{T^2}})$ $+\mathcal{O}(\frac{(\tau-1)^2}{T})+\mathcal{O}(\frac{\tau-1}{T^2})$~\cite{reisizadeh2020fedpaq}. In particular, any pick of $\tau = \mathcal{O}(\sqrt{T})$ ensures the convergence of the FL to the global optimal. For
smooth non-convex loss functions, the convergence rate is $ \mathcal{O}(\frac{1}{\sqrt{T}})+ \mathcal{O}(\frac{\tau-1}{T})$.

\subsubsection{Gradient sparsification}
Gradient sparsification methods~\cite{han2020adaptive,eghlidi2020sparse} convert a dense gradient into a sparse one by retaining only a subset of significant elements and setting the remaining coordinates to zero. Two commonly used techniques are rand-$k$ sparsification~\cite{eghlidi2020sparse} and top-$k$ sparsification~\cite{sattler2019robust,liu2020fedsel,lu2023top}. In particular, rand-$k$ sparsification randomly selects $k$ elements from the gradients, whereas top-$k$ sparsification retains the $k$ elements with the highest absolute values. It has been proved~\cite{wangni2018gradient} that rand-$k$ sparsification is an unbiased compression operator which, however, results in larger compression errors and therefore makes it less effective in practice compared to top-$k$ sparsification when high compression is required~\cite{shi2019understanding}.

Quantization~\cite{reisizadeh2020fedpaq,shlezinger2020uveqfed,kwon2023tighter} is another classic sparsification method that reduces the model size by lowering the bit width from 32-bit floating-point to a smaller precision. For now, numerous quantization methods (e.g., stochastic quantization~\cite{alistarh2017qsgd}, rotation-based quantization~\cite{suresh2017distributed}, etc.) have been proposed to compress model gradients of each client into a discrete set. Theoretical analysis shows that the convergence rate of FL with quantization has the order of $\mathcal{O}(\frac{1}{\sqrt{T}})$, which is the same as that of a classical FL framework for a non-convex loss function~\cite{reisizadeh2020fedpaq}.

Generally, top-$k$ sparsification methods can reduce communication complexity more efficiently compared to quantization. It has been demonstrated that the top-k sparsification with error feedback can accelerate convergence and accuracy with over 99\% gradient elements zeroed out~\cite{aji2017sparse}.

\subsection{Aggregation Rules}

The most widely used aggregation rule is FedAvg, which takes weighted average over all local gradients according to local data sizes. However, FedAvg implicitly assumes the equal quality of all gradients, limiting its effectiveness in scenarios like asynchronous FL (e.g., some gradients are stale~\cite{zhou2022towards} ) and Byzantine attack settings (e.g., some gradients are fake~\cite{liu2023byzantine}). To overcome these limitations, a variety of aggregation rules~\cite{zhu2023byzantine} have been proposed to optimize the model utility (e.g., model convergence and Byzantine robustness). Advanced aggregation methods can be roughly divided into two categories: weighting-based aggregation and statistic-based aggregation.

\subsubsection{Weighting-based aggregation}

Weighting-based aggregation rules aim to optimize the model performance (e.g., convergence, fairness, etc.) by assigning differentiated weights to local gradients according to specific statistical indicators (e.g., gradient staleness, model loss. etc.). The aggregation rule is as follows:
\begin{equation}
    \nabla f(x) = \frac{\sum_{i=1}^{N} a_{i} \nabla f(x_{i})}{\sum_{i=1}^{N} a_{i}},
\end{equation}
where $a_{i} \geq 0$ represents the weight for local gradient $\nabla f(x_{i})$. 

In asynchronous FL, with the objective of improving model convergence rate and decreasing errors, the weights are mainly determined by staleness~\cite{miao2023robust, chen2019communication} or descent direction~\cite{zhou2022towards}. Stale gradients typically exhibit biased descent directions and larger norms than normal gradients. Consequently, simply averaging the stale gradients would result in their dominance during the training process, leading to a sub-linear convergence rate of $\mathcal{O} (\frac{1}{\sqrt{T}})$ under the assumptions of general convexity~\cite{liu2015asynchronous} or bounded gradient~\cite{koloskova2022sharper}. Motivated by the intuition that low-staleness gradients are more reliable and accurate, it is reasonable to have the server take the weighted average over all gradients according to their staleness to estimate an unbiased gradient. Nonetheless, some high-staleness gradients also exhibit consistent descent directions to unbiased gradients, which can be utilized to accelerate model convergence~\cite{zhou2022towards,mitliagkas2016asynchrony}. Consequently, existing research enables the server to first evaluate the consistency of stale gradients with the estimated unbiased gradient, and accordingly assigns differentiated weights to those gradients to improve the convergence rate~\cite{zhou2022towards}. Through increasing the contributions of consistent gradients to the aggregation, such weighted aggregation method can significantly decrease the error in each round. It has been demonstrated that weighted asynchronous FL can achieve a convergence rate of $\mathcal{O} (\frac{1}{T})$ even under the assumption of non-convexity~\cite{zhou2022towards, dun2023efficient}.

In fair FL, to achieve performance consistency, most fair algorithms primarily enhance optimization objectives by amplifying the dominance of the large losses, which ultimately influences the aggregation weights in the training process~\cite{du2021fairness, cui2021addressing, li2021ditto}. The fundamental idea is to encourage the global model to demonstrate a favorable inclination towards disadvantaged clients (i.e. clients with smaller losses) by augmenting their weights. A straightforward method is to assign larger weights to gradients with larger loss values~\cite{cui2021addressing, du2021fairness}. Besides, several research introduces fairness constraints (e.g., the limited difference of predicted label probability distribution under different sensitive attributes~\cite{cui2021addressing}) 
to the FL optimization problem~\cite{cui2021addressing, du2021fairness}. To solve the constrained problem, the server first assigns different fairness budgets to clients and then adjusts the weights of different gradients according to whether the model performance inconsistency surpasses the assigned fairness budget.

\subsubsection{Statistic-based aggregation}

Statistic-based aggregation rules~\cite{blanchard2017machine,yin2018byzantine,guerraoui2018hidden,farhadkhani2022byzantine,karimireddy2021byzantine, zhu2023byzantine,liu2023byzantine} generate some robust statistics over the local gradients to perform aggregation, aiming to optimize the model utility with the presence of some fake gradients. The typical FedAvg, which takes the arithmetic mean of gradients to update the model, may diverge in the optimization process due to its vulnerability to outliers. In contrast, robust averaging methods~\cite{blanchard2017machine,yin2018byzantine,guerraoui2018hidden,farhadkhani2022byzantine} demonstrate improved robustness, i.e.,  median/trimmed-mean~\cite{yin2018byzantine} can achieve order-optimal error rates while maintaining a convergence rate of $O(\frac{\sigma}{\sqrt{KT}})$ under the assumption of non-convexity. However, these methods assume that honest gradients are close to each other and overlook data heterogeneity, making them vulnerable in non-i.i.d. FL settings. 

To address the impact of non-i.i.d. data on robustness, recent studies~\cite{karimireddy2021byzantine, zhu2023byzantine} leverages a bucketing step that groups heterogeneous gradients and computes the average within each bucket. This pre-processing step generates more homogeneous gradients, enabling robust aggregation and producing more resilient results. The bucketing-based methods have been shown to converge with a rate of $O(\frac{\sigma}{\sqrt{KT}})$, even under the assumptions of data heterogeneity and non-convexity.
Furthermore, a large dimension of model gradients can also amplify the effects of malicious gradients in non-i.i.d. data settings. To this end, GAS~\cite{liu2023byzantine} splits high-dimensional gradients into multiple low-dimensional subsets to mitigate these impacts. Robust statistic-based rules can then be applied to each subset, followed by concatenation. This splitting method is proven to be effective in reducing the impact of large gradients dimension $d$.

Existing studies~\cite{wu2020federated, gorbunovvariance} have demonstrated that the variance of stochastic gradients significantly impacts the Byzantine robustness of FL and a higher variance in stochastic gradients leads to weaker robustness. Hence, some research~\cite{wu2020federated, fedin2023byzantine, gorbunovvariance} combines existing variance reduction algorithms with robust statistics to mitigate the impact of gradient variances on Byzantine robustness. Specifically, Byra-SAGA~\cite{wu2020federated} proposes the integration of the distributed SAGA algorithm with the geometric median statistics achieving enhanced robustness. Theoretical results suggest that Byra-SAGA can achieve a convergence rate of $O(\frac{1}{T})$. Byz-VR-MARINA~\cite{gorbunovvariance} introduces the combined gradient compression algorithm VR-MARINA, variance reduction algorithm SARAH/PAGE, and robust aggregation rules, such as the geometric median statistics-based rule, to simultaneously achieve communication compression and robustness enhancement. Theoretical analysis indicates that under general non-convex assumptions, Byz-VR-MARINA can achieve a convergence rate of $O(\frac{1}{T})$.

Additionally, recent research~\cite{cheng2024momentum,su2015fault,fang2022bridge,yang2019byrdie,peng2021byzantine,he2022byzantine} has explored the extension of statistic-based aggregation rules to decentralized FL. For example, BRIDGE~\cite{fang2022bridge} extends the coordinate-wise trimmed-mean to decentralized FL. Theoretical findings indicate that, in a statistical sense, BRIDGE can converge at a convergence rate of $O(\frac{1}{T})$ to the optimal solution and a first-order stationary point in convex and non-convex settings respectively. IOS~\cite{wu2023byzantine} provides general guidelines to design Byzantine robust statistics for aggregation in decentralized FL, and proposes an effective method that iteratively discards models that are farthest away from the weighted average of models from neighboring nodes.

\section{Possible Future Directions}\label{sec:future_direction}

\subsection{Discussions for FL Development}

Along with extensive academic research, FL has also shown great potential and even practical applications in several industrial scenarios. For instance, Google has deployed FL on billions of Android systems to enable precise next-word prediction for mobile keyboards~\cite{hard2018federated}. NVIDIA has applied FL in medical image analysis across multiple institutions~\cite{roth2022nvidia}. As artificial intelligence has been becoming pervasive while privacy consciousness is ever increasing, FL is believed to thrive in the following future application scenarios.

\begin{itemize}
\item \textbf{FL in autonomous cars and robots.} Both self-driving cars and robots are expected to bloom soon, which will accumulate huge amounts of data in diverse environments. On the other hand, both applications are highly driven by various ML technologies, which require repeated training over massive data. At the same time, these applications may experience varying communication channels in the real world. Clearly, FL is very promising to be applied for accelerating their model learning while limiting data transfer and privacy disclosure~\cite{posner2021federated, savazzi2021opportunities}.

\item \textbf{Vertical federated Learning.} Besides these applications based on traditional horizontal FL~\cite{yang2019federated} (where the feature space is identical but the sample space is different), there are also explorations for vertical FL~\cite{liu2024vertical} (where the sample space is overlapped but the feature space is orthogonal). For example, Webank has applied vertical FL in financial risk controls by sharing knowledge between banks and insurance companies~\cite{yang2019federated, wu2020privacy}. Also, many internet companies like Bytedance has adopted vertical FL for intelligent recommendation in e-commence~\cite{li2024refer}. 

\item \textbf{FL for large models (LMs).} Large Models (LMs) have recently demonstrated astonishing AI abilities, gaining massive attention. To exploit the full potential of LMs, it often has to fine-tune them to domain-specific tasks or adapt them with domain-specific knowledge~\cite{wu2024fedbiot}. 
However, the fine-tuning of LMs not only requires rather powerful computing resources but also relies on a large amount of high-quality domain-specific data~\cite{kang2023grounding}, which however may be scattered among multiple sites and cannot be centralized. In this case, FL has been becoming a promising technology for achieving privacy-preserving LMs fine-tuning, thus truely grounding the large foundation models~\cite{cheng2021fine}.

\end{itemize}

\subsection{Future Directions for FL Optimization}
Despite the numerous advanced studies and promising applications, there still remains considerable space for the foundational FL optimization in terms of the optimization methods, and the practical system and privacy constraints.  
In the following, we discuss the possible future directions in FL optimization from three perspectives: optimization, system, and privacy.

\begin{itemize}

\item \textbf{New theories and methods for black-box optimization.} 
As we try to optimize the increasingly complicated AI systems, many practical learning tasks are essentially black-box optimization~\cite{guidotti2018survey,conn2009introduction,liu2020primer}, which can hardly give the analytic expression of the loss function or the gradient information. In such cases, besides zeroth-order optimization~\cite{conn2009introduction,liu2020primer} without the gradient information, we sometimes need more sophisticated optimization theories and methods. For example, hyper-parameter optimization~\cite{bergstra2011algorithms,bergstra2012random,yu2020hyper} and neural architecture search~\cite{white2021bananas,ren2021comprehensive} are very promising for optimizing the complex deep learning models. These problems are often modeled as Bi-level optimization where an optimization problem contains another optimization problem as a constraint~\cite{zhang2022bambi, li2024communication,huang2023achieving,tarzanagh2022fednest, yang2024simfbo, t2020personalized} in essence.  
However, how to adapt these advanced optimizations to the FL setting remains largely unexplored, thus being a promising research direction. 

\item \textbf{Optimization under practical FL system constraints.} FL optimization differs from conventional distributed optimization in many practical system constraints, including the underlying topology~\cite{ye2022decentralized, 9850408, warnat2021swarm}, parallel mode~\cite{doi:10.1137/1.9781611978032.95,nguyen2022federated}, resource limit~\cite{zhao2024faster,gauthier2022resource}, and data distribution~\cite{zhu2021federated, kairouz2021advances} etc. Existing studies have extensively explored the issue of non-i.i.d. data distribution, but still lack the deep consideration of many system constraints like topology and resource, especially their combinations. As discussed before, the underlying topologies can impact both the convergence rate and stability of FL optimization. In fact, the possible imbalance of resources across computing nodes, can impact not only the choice of the FL parallel mode but also the convergence performance. Therefore, future work may consider how to achieve faster and more accurate FL optimization by simultaneously optimizing the topology and resource allocation among FL nodes.

\item \textbf{FL optimization aware of privacy-preserving techniques.} FL optimization is proposed as a privacy-aware optimization method by design. Therefore, beyond DP, FL is often incorporated with many other privacy-preserving technologies like multi-party computation~\cite{keller2020mp,kadhe2020fastsecagg} and homomorphic encryption~\cite{acar2018survey,ma2022privacy}. However, similar to the degraded convergence incurred by DP constraints, the additional constraints in these privacy-preserving methods bring new challenges for FL optimization. For example, encryption-based methods often require quantization of exchanged gradient information to reduce the computing complexity~\cite{zhang2020batchcrypt,bonawitz2017practical,chen2022poisson}. This would further result in a degraded performance of FL optimization. Also, the cryptographic primitives significantly burden the computation and communication of FL optimization, which again complicates the trilemma among privacy, utility and efficiency. Therefore, an important question is how to design FL algorithms to optimize the tradeoff among the accuracy, privacy and complexity.
\end{itemize}

\section{Conclusion}\label{sec:conclusion}
 
As an important interdisciplinary research area in both applied mathematics and information sciences, FL still lacks a summarization of advanced studies in terms of mathematical optimization. To this end, we presented the first systematic survey on the assumptions, formulations, methods, and theoretical results in FL optimization, mainly focusing on the optimization challenges induced by non-i.i.d. data, rigorous privacy guarantee, decentralized topology, and online settings. Besides, we also reviewed some other important works on sparsification methods and aggregation rules, which can also improve FL optimization.
Finally, we envisioned the applications of FL in the AI era and} discussed several broader future directions from the perspectives of optimization, system, and privacy respectively.

\newpage

\bibliographystyle{unsrtnat}
\bibliography{csam}

\begin{thebibliography}{239}
\providecommand{\natexlab}[1]{#1}
\providecommand{\url}[1]{\texttt{#1}}
\expandafter\ifx\csname urlstyle\endcsname\relax
  \providecommand{\doi}[1]{doi: #1}\else
  \providecommand{\doi}{doi: \begingroup \urlstyle{rm}\Url}\fi

\bibitem[Voigt and Von~dem Bussche(2017)]{voigt2017eu}
Paul Voigt and Axel Von~dem Bussche.
\newblock The eu general data protection regulation (gdpr).
\newblock \emph{A Practical Guide, 1st Ed., Cham: Springer International
  Publishing}, 10\penalty0 (3152676):\penalty0 10--5555, 2017.

\bibitem[Pardau(2018)]{pardau2018california}
Stuart~L Pardau.
\newblock The california consumer privacy act: Towards a european-style privacy
  regime in the united states.
\newblock \emph{J. Tech. L. \& Pol'y}, 23:\penalty0 68, 2018.

\bibitem[Kalra et~al.(2023)Kalra, Wen, Cresswell, Volkovs, and
  Tizhoosh]{kalra2023decentralized}
Shivam Kalra, Junfeng Wen, Jesse~C Cresswell, Maksims Volkovs, and Hamid~R
  Tizhoosh.
\newblock Decentralized federated learning through proxy model sharing.
\newblock \emph{Nature communications}, 14\penalty0 (1):\penalty0 2899, 2023.

\bibitem[Hard et~al.(2018)Hard, Rao, Mathews, Ramaswamy, Beaufays, Augenstein,
  Eichner, Kiddon, and Ramage]{hard2018federated}
Andrew Hard, Kanishka Rao, Rajiv Mathews, Swaroop Ramaswamy, Fran{\c{c}}oise
  Beaufays, Sean Augenstein, Hubert Eichner, Chlo{\'e} Kiddon, and Daniel
  Ramage.
\newblock Federated learning for mobile keyboard prediction.
\newblock \emph{arXiv preprint arXiv:1811.03604}, 2018.

\bibitem[Kairouz et~al.(2021)Kairouz, McMahan, Avent, Bellet, Bennis, Bhagoji,
  Bonawitz, Charles, Cormode, Cummings, et~al.]{kairouz2021advances}
Peter Kairouz, H~Brendan McMahan, Brendan Avent, Aur{\'e}lien Bellet, Mehdi
  Bennis, Arjun~Nitin Bhagoji, Kallista Bonawitz, Zachary Charles, Graham
  Cormode, Rachel Cummings, et~al.
\newblock Advances and open problems in federated learning.
\newblock \emph{Foundations and trends{\textregistered} in machine learning},
  14\penalty0 (1--2):\penalty0 1--210, 2021.

\bibitem[Ogier~du Terrail et~al.(2023)Ogier~du Terrail, Leopold, Joly,
  B{\'e}guier, Andreux, Maussion, Schmauch, Tramel, Bendjebbar, Zaslavskiy,
  et~al.]{ogier2023federated}
Jean Ogier~du Terrail, Armand Leopold, Cl{\'e}ment Joly, Constance B{\'e}guier,
  Mathieu Andreux, Charles Maussion, Beno{\^\i}t Schmauch, Eric~W Tramel,
  Etienne Bendjebbar, Mikhail Zaslavskiy, et~al.
\newblock Federated learning for predicting histological response to
  neoadjuvant chemotherapy in triple-negative breast cancer.
\newblock \emph{Nature medicine}, 29\penalty0 (1):\penalty0 135--146, 2023.

\bibitem[Yang et~al.(2019{\natexlab{a}})Yang, Zhang, Ye, Li, and
  Xu]{yang2019ffd}
Wensi Yang, Yuhang Zhang, Kejiang Ye, Li~Li, and Cheng-Zhong Xu.
\newblock Ffd: A federated learning based method for credit card fraud
  detection.
\newblock In \emph{Proceedings of 8th International Congress on BIGDATA 2019,
  held as Part of the Services Conference Federation (SCF)}, pages 18--32.
  Springer, 2019{\natexlab{a}}.

\bibitem[Zhang et~al.(2024)Zhang, Li, Tian, Hao, and Zhang]{zhang2024vertical}
Rui Zhang, Hongwei Li, Luoding Tian, Meng Hao, and Yuan Zhang.
\newblock Vertical federated learning across heterogeneous regions for industry
  4.0.
\newblock \emph{IEEE Transactions on Industrial Informatics}, 2024.

\bibitem[Zhang et~al.(2023{\natexlab{a}})Zhang, Hong, and
  Elia]{zhang2023understanding}
Xinwei Zhang, Mingyi Hong, and Nicola Elia.
\newblock Understanding a class of decentralized and federated optimization
  algorithms: A multirate feedback control perspective.
\newblock \emph{SIAM Journal on Optimization}, 33\penalty0 (2):\penalty0
  652--683, 2023{\natexlab{a}}.

\bibitem[Touri and Gharesifard(2023)]{touri2023unified}
Behrouz Touri and Bahman Gharesifard.
\newblock A unified framework for continuous-time unconstrained distributed
  optimization.
\newblock \emph{SIAM Journal on Control and Optimization}, 61\penalty0
  (4):\penalty0 2004--2020, 2023.

\bibitem[Zhao et~al.(2024)Zhao, Burlachenko, Li, and
  Richt{\'a}rik]{zhao2024faster}
Haoyu Zhao, Konstantin Burlachenko, Zhize Li, and Peter Richt{\'a}rik.
\newblock Faster rates for compressed federated learning with client-variance
  reduction.
\newblock \emph{SIAM Journal on Mathematics of Data Science}, 6\penalty0
  (1):\penalty0 154--175, 2024.

\bibitem[Beznosikov et~al.(2023)Beznosikov, Horv{\'a}th, Richt{\'a}rik, and
  Safaryan]{beznosikov2023biased}
Aleksandr Beznosikov, Samuel Horv{\'a}th, Peter Richt{\'a}rik, and Mher
  Safaryan.
\newblock On biased compression for distributed learning.
\newblock \emph{Journal of Machine Learning Research}, 24\penalty0
  (276):\penalty0 1--50, 2023.

\bibitem[Lan and Zhou(2018)]{doi:10.1137/17M1157891}
Guanghui Lan and Yi~Zhou.
\newblock Random gradient extrapolation for distributed and stochastic
  optimization.
\newblock \emph{SIAM Journal on Optimization}, 28\penalty0 (4):\penalty0
  2753--2782, 2018.

\bibitem[Mishchenko et~al.(2020)Mishchenko, Iutzeler, and
  Malick]{doi:10.1137/18M1194699}
Konstantin Mishchenko, Franck Iutzeler, and J\'{e}r\^{o}me Malick.
\newblock A distributed flexible delay-tolerant proximal gradient algorithm.
\newblock \emph{SIAM Journal on Optimization}, 30\penalty0 (1):\penalty0
  933--959, 2020.

\bibitem[Beaude et~al.(2020)Beaude, Benchimol, Gaubert, Jacquot, and
  Oudjane]{doi:10.1137/19M127879X}
Olivier Beaude, Pascal Benchimol, St\'{e}phane Gaubert, Paulin Jacquot, and
  Nadia Oudjane.
\newblock A privacy-preserving method to optimize distributed resource
  allocation.
\newblock \emph{SIAM Journal on Optimization}, 30\penalty0 (3):\penalty0
  2303--2336, 2020.

\bibitem[Harks and Schwarz(2023)]{doi:10.1137/21M1400924}
Tobias Harks and Julian Schwarz.
\newblock A unified framework for pricing in nonconvex resource allocation
  games.
\newblock \emph{SIAM Journal on Optimization}, 33\penalty0 (2):\penalty0
  1223--1249, 2023.

\bibitem[Aybat and Hamedani(2019)]{doi:10.1137/17M1151973}
Necdet~Serhat Aybat and Erfan~Yazdandoost Hamedani.
\newblock A distributed admm-like method for resource sharing over time-varying
  networks.
\newblock \emph{SIAM Journal on Optimization}, 29\penalty0 (4):\penalty0
  3036--3068, 2019.

\bibitem[Zhu et~al.(2021)Zhu, Xu, Liu, and Jin]{zhu2021federated}
Hangyu Zhu, Jinjin Xu, Shiqing Liu, and Yaochu Jin.
\newblock Federated learning on non-iid data: A survey.
\newblock \emph{Neurocomputing}, 465:\penalty0 371--390, 2021.

\bibitem[Karimireddy et~al.(2020)Karimireddy, Kale, Mohri, Reddi, Stich, and
  Suresh]{karimireddy2020scaffold}
Sai~Praneeth Karimireddy, Satyen Kale, Mehryar Mohri, Sashank Reddi, Sebastian
  Stich, and Ananda~Theertha Suresh.
\newblock Scaffold: Stochastic controlled averaging for federated learning.
\newblock In \emph{International Conference on Machine Learning (ICML)}, pages
  5132--5143. PMLR, 2020.

\bibitem[Dwork et~al.(2006)Dwork, McSherry, Nissim, and
  Smith]{dwork2006calibrating}
Cynthia Dwork, Frank McSherry, Kobbi Nissim, and Adam Smith.
\newblock Calibrating noise to sensitivity in private data analysis.
\newblock In \emph{Theory of Cryptography: Third Theory of Cryptography
  Conference (TCC)}, pages 265--284. Springer, 2006.

\bibitem[Abadi et~al.(2016)Abadi, Chu, Goodfellow, McMahan, Mironov, Talwar,
  and Zhang]{abadi2016deep}
Martin Abadi, Andy Chu, Ian Goodfellow, H~Brendan McMahan, Ilya Mironov, Kunal
  Talwar, and Li~Zhang.
\newblock Deep learning with differential privacy.
\newblock In \emph{Proceedings of the 2016 ACM SIGSAC conference on computer
  and communications security (CCS)}, pages 308--318, 2016.

\bibitem[Wu et~al.(2020{\natexlab{a}})Wu, Farokhi, Smith, and
  Kaafar]{wu2020value}
Nan Wu, Farhad Farokhi, David Smith, and Mohamed~Ali Kaafar.
\newblock The value of collaboration in convex machine learning with
  differential privacy.
\newblock In \emph{2020 IEEE Symposium on Security and Privacy (SP)}, pages
  304--317. IEEE, 2020{\natexlab{a}}.

\bibitem[Zhang et~al.(2022{\natexlab{a}})Zhang, Chen, Hong, Wu, and
  Yi]{zhang2022understanding}
Xinwei Zhang, Xiangyi Chen, Mingyi Hong, Zhiwei~Steven Wu, and Jinfeng Yi.
\newblock Understanding clipping for federated learning: Convergence and
  client-level differential privacy.
\newblock In \emph{International Conference on Machine Learning (ICML)},
  2022{\natexlab{a}}.

\bibitem[Hu et~al.(2023)Hu, Guo, and Gong]{hu2023federated}
Rui Hu, Yuanxiong Guo, and Yanmin Gong.
\newblock Federated learning with sparsified model perturbation: Improving
  accuracy under client-level differential privacy.
\newblock \emph{IEEE Transactions on Mobile Computing}, 2023.

\bibitem[Girgis et~al.(2021)Girgis, Data, Diggavi, Kairouz, and
  Suresh]{girgis2021shuffled}
Antonious Girgis, Deepesh Data, Suhas Diggavi, Peter Kairouz, and
  Ananda~Theertha Suresh.
\newblock Shuffled model of differential privacy in federated learning.
\newblock In \emph{International Conference on Artificial Intelligence and
  Statistics (AISTATS)}, pages 2521--2529. PMLR, 2021.

\bibitem[Hua et~al.(2022)Hua, Miller, Bertozzi, Qian, and
  Wang]{doi:10.1137/21M1465081}
Yifan Hua, Kevin Miller, Andrea~L. Bertozzi, Chen Qian, and Bao Wang.
\newblock Efficient and reliable overlay networks for decentralized federated
  learning.
\newblock \emph{SIAM Journal on Applied Mathematics}, 82\penalty0 (4):\penalty0
  1558--1586, 2022.

\bibitem[Yuan et~al.(2016)Yuan, Ling, and Yin]{yuan2016convergence}
Kun Yuan, Qing Ling, and Wotao Yin.
\newblock On the convergence of decentralized gradient descent.
\newblock \emph{SIAM Journal on Optimization}, 26\penalty0 (3):\penalty0
  1835--1854, 2016.

\bibitem[Hong and Chae(2021)]{hong2021communication}
Songnam Hong and Jeongmin Chae.
\newblock Communication-efficient randomized algorithm for multi-kernel online
  federated learning.
\newblock \emph{IEEE Transactions on pattern analysis and machine
  intelligence}, 44\penalty0 (12):\penalty0 9872--9886, 2021.

\bibitem[Chen et~al.(2020{\natexlab{a}})Chen, Ning, Slawski, and
  Rangwala]{chen2020asynchronous}
Yujing Chen, Yue Ning, Martin Slawski, and Huzefa Rangwala.
\newblock Asynchronous online federated learning for edge devices with non-iid
  data.
\newblock In \emph{2020 IEEE International Conference on Big Data (Big Data)},
  pages 15--24. IEEE, 2020{\natexlab{a}}.

\bibitem[M~Ghari and Shen(2022)]{m2022personalized}
Pouya M~Ghari and Yanning Shen.
\newblock Personalized online federated learning with multiple kernels.
\newblock In \emph{Advances in Neural Information Processing Systems
  (NeurIPS)}, volume~35, pages 33316--33329, 2022.

\bibitem[Kwon et~al.(2023)Kwon, Park, and Hong]{kwon2023tighter}
Dohyeok Kwon, Jonghwan Park, and Songnam Hong.
\newblock Tighter regret analysis and optimization of online federated
  learning.
\newblock \emph{IEEE Transactions on Pattern Analysis and Machine
  Intelligence}, 2023.

\bibitem[Gauthier et~al.(2022)Gauthier, Gogineni, Werner, Huang, and
  Kuh]{gauthier2022resource}
Francois Gauthier, Vinay~Chakravarthi Gogineni, Stefan Werner, Yih-Fang Huang,
  and Anthony Kuh.
\newblock Resource-aware asynchronous online federated learning for nonlinear
  regression.
\newblock In \emph{ICC 2022-IEEE International Conference on Communications
  (ICC)}, pages 2828--2833. IEEE, 2022.

\bibitem[Antunes et~al.(2022)Antunes, Andr{\'e}~da Costa, K{\"u}derle, Yari,
  and Eskofier]{antunes2022federated}
Rodolfo~Stoffel Antunes, Cristiano Andr{\'e}~da Costa, Arne K{\"u}derle,
  Imrana~Abdullahi Yari, and Bj{\"o}rn Eskofier.
\newblock Federated learning for healthcare: Systematic review and architecture
  proposal.
\newblock \emph{ACM Transactions on Intelligent Systems and Technology (TIST)},
  13\penalty0 (4):\penalty0 1--23, 2022.

\bibitem[Lu et~al.(2024)Lu, Pan, Dai, Si, and Zhang]{lu2024federated}
Zili Lu, Heng Pan, Yueyue Dai, Xueming Si, and Yan Zhang.
\newblock Federated learning with non-iid data: A survey.
\newblock \emph{IEEE Internet of Things Journal}, 2024.

\bibitem[Yang et~al.(2019{\natexlab{b}})Yang, Liu, Chen, and
  Tong]{yang2019federated}
Qiang Yang, Yang Liu, Tianjian Chen, and Yongxin Tong.
\newblock Federated machine learning: Concept and applications.
\newblock \emph{ACM Transactions on Intelligent Systems and Technology (TIST)},
  10\penalty0 (2):\penalty0 1--19, 2019{\natexlab{b}}.

\bibitem[Ren et~al.(2024)Ren, Yang, Zhao, McCann, and Xu]{ren2024belt}
Xuebin Ren, Shusen Yang, Cong Zhao, Julie McCann, and Zongben Xu.
\newblock Belt and braces: When federated learning meets differential privacy.
\newblock \emph{Communications of the ACM (CACM)}, 67\penalty0 (12):\penalty0
  66–77, 2024.

\bibitem[Kumar et~al.(2023)Kumar, Mohan, and Cenkeramaddi]{kumar2023impact}
K~Naveen Kumar, C~Krishna Mohan, and Linga~Reddy Cenkeramaddi.
\newblock The impact of adversarial attacks on federated learning: A survey.
\newblock \emph{IEEE Transactions on Pattern Analysis and Machine
  Intelligence}, 2023.

\bibitem[Nguyen et~al.(2022)Nguyen, Pham, Pathirana, Ding, Seneviratne, Lin,
  Dobre, and Hwang]{nguyen2022federated}
Dinh~C Nguyen, Quoc-Viet Pham, Pubudu~N Pathirana, Ming Ding, Aruna
  Seneviratne, Zihuai Lin, Octavia Dobre, and Won-Joo Hwang.
\newblock Federated learning for smart healthcare: A survey.
\newblock \emph{ACM Computing Surveys (Csur)}, 55\penalty0 (3):\penalty0 1--37,
  2022.

\bibitem[McMahan et~al.(2017)McMahan, Moore, Ramage, Hampson, and
  y~Arcas]{mcmahan2017communication}
Brendan McMahan, Eider Moore, Daniel Ramage, Seth Hampson, and Blaise~Aguera
  y~Arcas.
\newblock Communication-efficient learning of deep networks from decentralized
  data.
\newblock In \emph{International Conference on Artificial Intelligence and
  Statistics (AISTATS)}, pages 1273--1282. PMLR, 2017.

\bibitem[Liu et~al.(2024)Liu, Kang, Zou, Pu, He, Ye, Ouyang, Zhang, and
  Yang]{liu2024vertical}
Yang Liu, Yan Kang, Tianyuan Zou, Yanhong Pu, Yuanqin He, Xiaozhou Ye,
  Ye~Ouyang, Ya-Qin Zhang, and Qiang Yang.
\newblock Vertical federated learning: Concepts, advances, and challenges.
\newblock \emph{IEEE Transactions on Knowledge and Data Engineering}, 2024.

\bibitem[Jin et~al.(2021)Jin, Chen, Hsu, Yu, and Chen]{jin2021cafe}
Xiao Jin, Pin-Yu Chen, Chia-Yi Hsu, Chia-Mu Yu, and Tianyi Chen.
\newblock Cafe: Catastrophic data leakage in vertical federated learning.
\newblock In \emph{Advances in Neural Information Processing Systems
  (NeurIPS)}, volume~34, pages 994--1006, 2021.

\bibitem[Wang et~al.(2024{\natexlab{a}})Wang, Gu, Zhang, Li, Wang, and
  Ling]{wang2024unified}
Ganyu Wang, Bin Gu, Qingsong Zhang, Xiang Li, Boyu Wang, and Charles~X Ling.
\newblock A unified solution for privacy and communication efficiency in
  vertical federated learning.
\newblock In \emph{Advances in Neural Information Processing Systems
  (NeurIPS)}, volume~36, 2024{\natexlab{a}}.

\bibitem[Bubeck et~al.(2015)]{bubeck2015convex}
S{\'e}bastien Bubeck et~al.
\newblock Convex optimization: Algorithms and complexity.
\newblock \emph{Foundations and Trends{\textregistered} in Machine Learning},
  8\penalty0 (3-4):\penalty0 231--357, 2015.

\bibitem[Reddi et~al.(2020)Reddi, Charles, Zaheer, Garrett, Rush,
  Kone{\v{c}}n{\`y}, Kumar, and McMahan]{reddi2020adaptive}
Sashank~J Reddi, Zachary Charles, Manzil Zaheer, Zachary Garrett, Keith Rush,
  Jakub Kone{\v{c}}n{\`y}, Sanjiv Kumar, and Hugh~Brendan McMahan.
\newblock Adaptive federated optimization.
\newblock In \emph{International Conference on Learning Representations
  (ICLR)}, 2020.

\bibitem[Stich and Karimireddy(2019)]{stich2019error}
Sebastian~U Stich and Sai~Praneeth Karimireddy.
\newblock The error-feedback framework: Better rates for sgd with delayed
  gradients and compressed communication.
\newblock \emph{arXiv preprint arXiv:1909.05350}, 2019.

\bibitem[Yu et~al.(2018)Yu, Yang, and Zhu]{yu2018parallel}
Hao Yu, Sen Yang, and Shenghuo Zhu.
\newblock Parallel restarted sgd for non-convex optimization with faster
  convergence and less communication.
\newblock \emph{arXiv preprint arXiv:1807.06629}, 2\penalty0 (4):\penalty0 7,
  2018.

\bibitem[Karimireddy et~al.(2021{\natexlab{a}})Karimireddy, He, and
  Jaggi]{karimireddy2021learning}
Sai~Praneeth Karimireddy, Lie He, and Martin Jaggi.
\newblock Learning from history for byzantine robust optimization.
\newblock In \emph{International Conference on Machine Learning (ICML)}, pages
  5311--5319. PMLR, 2021{\natexlab{a}}.

\bibitem[Yang et~al.(2022)Yang, Bao, Yuan, Tran, and Zomaya]{yang2022federated}
Zhengjie Yang, Wei Bao, Dong Yuan, Nguyen~H Tran, and Albert~Y Zomaya.
\newblock Federated learning with nesterov accelerated gradient.
\newblock \emph{IEEE Transactions on Parallel and Distributed Systems},
  33\penalty0 (12):\penalty0 4863--4873, 2022.

\bibitem[Ju et~al.(2023)Ju, Zhang, Toor, and Hellander]{ju2023accelerating}
Li~Ju, Tianru Zhang, Salman Toor, and Andreas Hellander.
\newblock Accelerating fair federated learning: Adaptive federated adam.
\newblock \emph{arXiv preprint arXiv:2301.09357}, 2023.

\bibitem[Zhou and Li(2023)]{zhou2023federated}
Shenglong Zhou and Geoffrey~Ye Li.
\newblock Federated learning via inexact admm.
\newblock \emph{IEEE Transactions on Pattern Analysis and Machine
  Intelligence}, 2023.

\bibitem[Jeon et~al.(2021)Jeon, Ferdous, Rahman, and Walid]{jeon2021privacy}
Beomyeol Jeon, SM~Ferdous, Muntasir~Raihan Rahman, and Anwar Walid.
\newblock Privacy-preserving decentralized aggregation for federated learning.
\newblock In \emph{IEEE INFOCOM 2021-IEEE Conference on Computer Communications
  Workshops (INFOCOM WKSHPS)}, pages 1--6. IEEE, 2021.

\bibitem[Kant et~al.(2022)Kant, da~Silva, Fodor, G{\"o}ransson, Bengtsson, and
  Fischione]{kant2022federated}
Shashi Kant, Jos{\'e} Mairton~B da~Silva, Gabor Fodor, Bo~G{\"o}ransson, Mats
  Bengtsson, and Carlo Fischione.
\newblock Federated learning using three-operator admm.
\newblock \emph{IEEE Journal of Selected Topics in Signal Processing},
  17\penalty0 (1):\penalty0 205--221, 2022.

\bibitem[Elgabli et~al.(2022)Elgabli, Issaid, Bedi, Rajawat, Bennis, and
  Aggarwal]{elgabli2022fednew}
Anis Elgabli, Chaouki~Ben Issaid, Amrit~Singh Bedi, Ketan Rajawat, Mehdi
  Bennis, and Vaneet Aggarwal.
\newblock Fednew: A communication-efficient and privacy-preserving newton-type
  method for federated learning.
\newblock In \emph{International Conference on Machine Learning (ICML)}, pages
  5861--5877. PMLR, 2022.

\bibitem[Safaryan et~al.(2021)Safaryan, Islamov, Qian, and
  Richt{\'a}rik]{safaryan2021fednl}
Mher Safaryan, Rustem Islamov, Xun Qian, and Peter Richt{\'a}rik.
\newblock Fednl: Making newton-type methods applicable to federated learning.
\newblock \emph{arXiv preprint arXiv:2106.02969}, 2021.

\bibitem[Dinh et~al.(2022)Dinh, Tran, Nguyen, Bao, Balef, Zhou, and
  Zomaya]{dinh2022done}
Canh~T Dinh, Nguyen~H Tran, Tuan~Dung Nguyen, Wei Bao, Amir~Rezaei Balef,
  Bing~B Zhou, and Albert~Y Zomaya.
\newblock Done: distributed approximate newton-type method for federated edge
  learning.
\newblock \emph{IEEE Transactions on Parallel and Distributed Systems},
  33\penalty0 (11):\penalty0 2648--2660, 2022.

\bibitem[Nagaraju et~al.(2023)Nagaraju, Sen, and Mohan]{nagaraju2023fonn}
C~Nagaraju, Mrinmay Sen, and C~Krishna Mohan.
\newblock Fonn: Federated optimization with nys-newton.
\newblock In \emph{TENCON 2023-2023 IEEE Region 10 Conference (TENCON)}, pages
  530--534. IEEE, 2023.

\bibitem[Fang et~al.(2022{\natexlab{a}})Fang, Yu, Jiang, Shi, Jones, and
  Zhou]{fang2022communication}
Wenzhi Fang, Ziyi Yu, Yuning Jiang, Yuanming Shi, Colin~N Jones, and Yong Zhou.
\newblock Communication-efficient stochastic zeroth-order optimization for
  federated learning.
\newblock \emph{IEEE Transactions on Signal Processing}, 70:\penalty0
  5058--5073, 2022{\natexlab{a}}.

\bibitem[Haddadpour et~al.(2019)Haddadpour, Kamani, Mahdavi, and
  Cadambe]{haddadpour2019local}
Farzin Haddadpour, Mohammad~Mahdi Kamani, Mehrdad Mahdavi, and Viveck Cadambe.
\newblock Local sgd with periodic averaging: Tighter analysis and adaptive
  synchronization.
\newblock In \emph{Advances in Neural Information Processing Systems
  (NeurIPS)}, volume~32, 2019.

\bibitem[Qiu et~al.(2023)Qiu, Shanbhag, and Yousefian]{qiu2023zeroth}
Yuyang Qiu, Uday Shanbhag, and Farzad Yousefian.
\newblock Zeroth-order methods for nondifferentiable, nonconvex, and
  hierarchical federated optimization.
\newblock In \emph{Advances in Neural Information Processing Systems
  (NeurIPS)}, volume~36, 2023.

\bibitem[Beck(2017)]{beck2017first}
Amir Beck.
\newblock \emph{First-order methods in optimization}.
\newblock SIAM, 2017.

\bibitem[Bottou(2010)]{bottou2010large}
L{\'e}on Bottou.
\newblock Large-scale machine learning with stochastic gradient descent.
\newblock In \emph{19th International Conference on Computational
  StatisticsParis France (COMPSTAT)}, pages 177--186. Springer, 2010.

\bibitem[Li et~al.(2019)Li, Huang, Yang, Wang, and Zhang]{li2019convergence}
Xiang Li, Kaixuan Huang, Wenhao Yang, Shusen Wang, and Zhihua Zhang.
\newblock On the convergence of fedavg on non-iid data.
\newblock In \emph{International Conference on Learning Representations
  (ICLR)}, 2019.

\bibitem[Li et~al.(2020)Li, Sahu, Zaheer, Sanjabi, Talwalkar, and
  Smith]{li2020federated}
Tian Li, Anit~Kumar Sahu, Manzil Zaheer, Maziar Sanjabi, Ameet Talwalkar, and
  Virginia Smith.
\newblock Federated optimization in heterogeneous networks.
\newblock In \emph{Proceedings of Machine learning and systems (MLSys)},
  volume~2, pages 429--450, 2020.

\bibitem[Liu et~al.(2020{\natexlab{a}})Liu, Chen, Chen, and
  Zhang]{liu2020accelerating}
Wei Liu, Li~Chen, Yunfei Chen, and Wenyi Zhang.
\newblock Accelerating federated learning via momentum gradient descent.
\newblock \emph{IEEE Transactions on Parallel and Distributed Systems},
  31\penalty0 (8):\penalty0 1754--1766, 2020{\natexlab{a}}.

\bibitem[Huo et~al.(2020)Huo, Yang, Gu, Huang, et~al.]{huo2020faster}
Zhouyuan Huo, Qian Yang, Bin Gu, Lawrence~Carin Huang, et~al.
\newblock Faster on-device training using new federated momentum algorithm.
\newblock \emph{arXiv preprint arXiv:2002.02090}, 2020.

\bibitem[Xu et~al.(2021)Xu, Wang, Wang, and Yao]{xu2021fedcm}
Jing Xu, Sen Wang, Liwei Wang, and Andrew Chi-Chih Yao.
\newblock Fedcm: Federated learning with client-level momentum.
\newblock \emph{arXiv preprint arXiv:2106.10874}, 2021.

\bibitem[Sun et~al.(2024)Sun, Wu, Huang, and Zhang]{sun2024role}
Jianhui Sun, Xidong Wu, Heng Huang, and Aidong Zhang.
\newblock On the role of server momentum in federated learning.
\newblock In \emph{Proceedings of the AAAI Conference on Artificial
  Intelligence (AAAI)}, volume~38, pages 15164--15172, 2024.

\bibitem[Hinton et~al.(2012)Hinton, Srivastava, and Swersky]{hinton2012neural}
Geoffrey Hinton, Nitish Srivastava, and Kevin Swersky.
\newblock Neural networks for machine learning lecture 6a overview of
  mini-batch gradient descent.
\newblock \emph{Cited on}, 14\penalty0 (8):\penalty0 2, 2012.

\bibitem[Liang et~al.(2019)Liang, Shen, Liu, Pan, Chen, and
  Cheng]{liang2019variance}
Xianfeng Liang, Shuheng Shen, Jingchang Liu, Zhen Pan, Enhong Chen, and Yifei
  Cheng.
\newblock Variance reduced local sgd with lower communication complexity.
\newblock \emph{arXiv preprint arXiv:1912.12844}, 2019.

\bibitem[Murata and Suzuki(2021)]{murata2021bias}
Tomoya Murata and Taiji Suzuki.
\newblock Bias-variance reduced local sgd for less heterogeneous federated
  learning.
\newblock \emph{arXiv preprint arXiv:2102.03198}, 2021.

\bibitem[Gao et~al.(2022)Gao, Fu, Li, Chen, Xu, and Xu]{gao2022feddc}
Liang Gao, Huazhu Fu, Li~Li, Yingwen Chen, Ming Xu, and Cheng-Zhong Xu.
\newblock Feddc: Federated learning with non-iid data via local drift
  decoupling and correction.
\newblock In \emph{Proceedings of the IEEE/CVF Conference on Computer Vision
  and Pattern Recognition (CVPR)}, pages 10112--10121, 2022.

\bibitem[Agarwal et~al.(2017)Agarwal, Bullins, and Hazan]{agarwal2017second}
Naman Agarwal, Brian Bullins, and Elad Hazan.
\newblock Second-order stochastic optimization for machine learning in linear
  time.
\newblock \emph{Journal of Machine Learning Research}, 18\penalty0
  (116):\penalty0 1--40, 2017.

\bibitem[Sun et~al.(2019)Sun, Cao, Zhu, and Zhao]{sun2019survey}
Shiliang Sun, Zehui Cao, Han Zhu, and Jing Zhao.
\newblock A survey of optimization methods from a machine learning perspective.
\newblock \emph{IEEE transactions on cybernetics}, 50\penalty0 (8):\penalty0
  3668--3681, 2019.

\bibitem[Boyd and Vandenberghe(2004)]{boyd2004convex}
Stephen~P Boyd and Lieven Vandenberghe.
\newblock \emph{Convex optimization}.
\newblock Cambridge university press, 2004.

\bibitem[Guidotti et~al.(2018)Guidotti, Monreale, Ruggieri, Turini, Giannotti,
  and Pedreschi]{guidotti2018survey}
Riccardo Guidotti, Anna Monreale, Salvatore Ruggieri, Franco Turini, Fosca
  Giannotti, and Dino Pedreschi.
\newblock A survey of methods for explaining black box models.
\newblock \emph{ACM computing surveys (CSUR)}, 51\penalty0 (5):\penalty0 1--42,
  2018.

\bibitem[Mania et~al.(2018)Mania, Guy, and Recht]{mania2018simple}
Horia Mania, Aurelia Guy, and Benjamin Recht.
\newblock Simple random search of static linear policies is competitive for
  reinforcement learning.
\newblock In \emph{Advances in Neural Information Processing Systems
  (NeurIPS)}, volume~31, 2018.

\bibitem[Zhan et~al.(2023)Zhan, Cen, Huang, Chen, Lee, and Chi]{zhan2023policy}
Wenhao Zhan, Shicong Cen, Baihe Huang, Yuxin Chen, Jason~D Lee, and Yuejie Chi.
\newblock Policy mirror descent for regularized reinforcement learning: A
  generalized framework with linear convergence.
\newblock \emph{SIAM Journal on Optimization}, 33\penalty0 (2):\penalty0
  1061--1091, 2023.

\bibitem[Conn et~al.(2009)Conn, Scheinberg, and Vicente]{conn2009introduction}
Andrew~R Conn, Katya Scheinberg, and Luis~N Vicente.
\newblock \emph{Introduction to derivative-free optimization}.
\newblock SIAM, 2009.

\bibitem[Liu et~al.(2020{\natexlab{b}})Liu, Chen, Kailkhura, Zhang, Hero~III,
  and Varshney]{liu2020primer}
Sijia Liu, Pin-Yu Chen, Bhavya Kailkhura, Gaoyuan Zhang, Alfred~O Hero~III, and
  Pramod~K Varshney.
\newblock A primer on zeroth-order optimization in signal processing and
  machine learning: Principals, recent advances, and applications.
\newblock \emph{IEEE Signal Processing Magazine}, 37\penalty0 (5):\penalty0
  43--54, 2020{\natexlab{b}}.

\bibitem[Chen et~al.(2024{\natexlab{a}})Chen, Chen, Gu, and Deng]{chen2024fine}
Jun Chen, Hong Chen, Bin Gu, and Hao Deng.
\newblock Fine-grained theoretical analysis of federated zeroth-order
  optimization.
\newblock In \emph{Advances in Neural Information Processing Systems
  (NeurIPS)}, volume~36, 2024{\natexlab{a}}.

\bibitem[Qian et~al.(2022)Qian, Islamov, Safaryan, and
  Richtarik]{qian2022basis}
Xun Qian, Rustem Islamov, Mher Safaryan, and Peter Richtarik.
\newblock Basis matters: Better communication-efficient second order methods
  for federated learning.
\newblock In \emph{International Conference on Artificial Intelligence and
  Statistics (AISTATS)}, pages 680--720. PMLR, 2022.

\bibitem[Stich(2018)]{Stich_2018}
Sebastian~U Stich.
\newblock Local sgd converges fast and communicates little.
\newblock \emph{arXiv preprint arXiv:1805.09767}, 2018.

\bibitem[Khaled et~al.(2020)Khaled, Mishchenko, and
  Richt{\'a}rik]{Khaled_Mishchenko_2020}
Ahmed Khaled, Konstantin Mishchenko, and Peter Richt{\'a}rik.
\newblock Tighter theory for local sgd on identical and heterogeneous data.
\newblock In \emph{International Conference on Artificial Intelligence and
  Statistics (AISTATS)}, pages 4519--4529. PMLR, 2020.

\bibitem[Chen et~al.(2024{\natexlab{b}})Chen, Li, and Chi]{chen2024escaping}
Sijin Chen, Zhize Li, and Yuejie Chi.
\newblock Escaping saddle points in heterogeneous federated learning via
  distributed sgd with communication compression.
\newblock In \emph{International Conference on Artificial Intelligence and
  Statistics (AISTATS)}, pages 2701--2709. PMLR, 2024{\natexlab{b}}.

\bibitem[Yu et~al.(2019)Yu, Yang, and Zhu]{yu2019parallel}
Hao Yu, Sen Yang, and Shenghuo Zhu.
\newblock Parallel restarted sgd with faster convergence and less
  communication: Demystifying why model averaging works for deep learning.
\newblock In \emph{Proceedings of the AAAI Conference on Artificial
  Intelligence (AAAI)}, volume~33, pages 5693--5700, 2019.

\bibitem[Wang et~al.(2020{\natexlab{a}})Wang, Liu, Liang, Joshi, and
  Poor]{wang2020tackling}
Jianyu Wang, Qinghua Liu, Hao Liang, Gauri Joshi, and H~Vincent Poor.
\newblock Tackling the objective inconsistency problem in heterogeneous
  federated optimization.
\newblock In \emph{Advances in Neural Information Processing Systems
  (NeurIPS)}, volume~33, pages 7611--7623, 2020{\natexlab{a}}.

\bibitem[Zhao et~al.(2018)Zhao, Li, Lai, Suda, Civin, and
  Chandra]{zhao2018federated}
Yue Zhao, Meng Li, Liangzhen Lai, Naveen Suda, Damon Civin, and Vikas Chandra.
\newblock Federated learning with non-iid data.
\newblock \emph{arXiv preprint arXiv:1806.00582}, 2018.

\bibitem[Shoham et~al.(2019)Shoham, Avidor, Keren, Israel, Benditkis,
  Mor-Yosef, and Zeitak]{shoham2019overcoming}
Neta Shoham, Tomer Avidor, Aviv Keren, Nadav Israel, Daniel Benditkis, Liron
  Mor-Yosef, and Itai Zeitak.
\newblock Overcoming forgetting in federated learning on non-iid data.
\newblock \emph{arXiv preprint arXiv:1910.07796}, 2019.

\bibitem[Li et~al.(2021)Li, Hu, Beirami, and Smith]{li2021ditto}
Tian Li, Shengyuan Hu, Ahmad Beirami, and Virginia Smith.
\newblock Ditto: Fair and robust federated learning through personalization.
\newblock In \emph{International Conference on Machine Learning (ICML)}, pages
  6357--6368. PMLR, 2021.

\bibitem[T~Dinh et~al.(2020)T~Dinh, Tran, and Nguyen]{t2020personalized}
Canh T~Dinh, Nguyen Tran, and Josh Nguyen.
\newblock Personalized federated learning with moreau envelopes.
\newblock In \emph{Advances in Neural Information Processing Systems
  (NeurIPS)}, volume~33, pages 21394--21405, 2020.

\bibitem[Mansour et~al.(2020)Mansour, Mohri, Ro, and Suresh]{mansour2020three}
Yishay Mansour, Mehryar Mohri, Jae Ro, and Ananda~Theertha Suresh.
\newblock Three approaches for personalization with applications to federated
  learning.
\newblock \emph{arXiv preprint arXiv:2002.10619}, 2020.

\bibitem[Sun et~al.(2023{\natexlab{a}})Sun, Shen, Chen, Ding, and
  Tao]{sun2023dynamic}
Yan Sun, Li~Shen, Shixiang Chen, Liang Ding, and Dacheng Tao.
\newblock Dynamic regularized sharpness aware minimization in federated
  learning: Approaching global consistency and smooth landscape.
\newblock In \emph{International Conference on Machine Learning (ICML)}, pages
  32991--33013. PMLR, 2023{\natexlab{a}}.

\bibitem[Yang et~al.(2023)Yang, Liu, Zhang, and Zhou]{yang2023personalized}
Zhikai Yang, Yaping Liu, Shuo Zhang, and Keshen Zhou.
\newblock Personalized federated learning with model interpolation among client
  clusters and its application in smart home.
\newblock In \emph{World Wide Web (WWW)}, volume~26, pages 2175--2200.
  Springer, 2023.

\bibitem[Chen et~al.(2023)Chen, Jiang, Dou, Wang, and Li]{chen2023fedsoup}
Minghui Chen, Meirui Jiang, Qi~Dou, Zehua Wang, and Xiaoxiao Li.
\newblock Fedsoup: Improving generalization and personalization in federated
  learning via selective model interpolation.
\newblock In \emph{International Conference on Medical Image Computing and
  Computer-Assisted Intervention (MICCAI)}, pages 318--328. Springer, 2023.

\bibitem[Ward et~al.(2020)Ward, Wu, and Bottou]{ward2020adagrad}
Rachel Ward, Xiaoxia Wu, and Leon Bottou.
\newblock Adagrad stepsizes: Sharp convergence over nonconvex landscapes.
\newblock \emph{Journal of Machine Learning Research}, 21\penalty0
  (219):\penalty0 1--30, 2020.

\bibitem[Zaheer et~al.(2018)Zaheer, Reddi, Sachan, Kale, and
  Kumar]{zaheer2018adaptive}
Manzil Zaheer, Sashank Reddi, Devendra Sachan, Satyen Kale, and Sanjiv Kumar.
\newblock Adaptive methods for nonconvex optimization.
\newblock In \emph{Advances in Neural Information Processing Systems
  (NeurIPS)}, volume~31, 2018.

\bibitem[Wang and Joshi(2018)]{wang2018cooperative}
Jianyu Wang and Gauri Joshi.
\newblock Cooperative sgd: A unified framework for the design and analysis of
  communication-efficient sgd algorithms.
\newblock \emph{arXiv preprint arXiv:1808.07576}, 2018.

\bibitem[Cheng et~al.(2024)Cheng, Huang, Wu, and Yuan]{cheng2024momentum}
Ziheng Cheng, Xinmeng Huang, Pengfei Wu, and Kun Yuan.
\newblock Momentum benefits non-iid federated learning simply and provably.
\newblock In \emph{International Conference on Learning Representations
  (ICLR)}, 2024.

\bibitem[Wang et~al.(2020{\natexlab{b}})Wang, Tantia, Ballas, and
  Rabbat]{Wang2020SlowMo}
Jianyu Wang, Vinayak Tantia, Nicolas Ballas, and Michael Rabbat.
\newblock Slowmo: Improving communication-efficient distributed sgd with slow
  momentum.
\newblock In \emph{International Conference on Learning Representations
  (ICLR)}, 2020{\natexlab{b}}.
\newblock URL \url{https://openreview.net/forum?id=SkxJ8REYPH}.

\bibitem[Ye et~al.(2023)Ye, Xu, Wang, Xu, Chen, and Wang]{ye2023feddisco}
Rui Ye, Mingkai Xu, Jianyu Wang, Chenxin Xu, Siheng Chen, and Yanfeng Wang.
\newblock Feddisco: Federated learning with discrepancy-aware collaboration.
\newblock In \emph{International Conference on Machine Learning (ICML)}, pages
  39879--39902. PMLR, 2023.

\bibitem[Ullman(2018)]{ullman2018tight}
Jonathan Ullman.
\newblock Tight lower bounds for locally differentially private selection.
\newblock \emph{arXiv preprint arXiv:1802.02638}, 2018.

\bibitem[Kuru et~al.(2022)Kuru, Ilker~Birbil, Gurbuzbalaban, and
  Yildirim]{kuru2022differentially}
Nurdan Kuru, S~Ilker~Birbil, Mert Gurbuzbalaban, and Sinan Yildirim.
\newblock Differentially private accelerated optimization algorithms.
\newblock \emph{SIAM Journal on Optimization}, 32\penalty0 (2):\penalty0
  795--821, 2022.

\bibitem[Chen and Wang(2024)]{chen2024locally}
Ziqin Chen and Yongqiang Wang.
\newblock Locally differentially private decentralized stochastic bilevel
  optimization with guaranteed convergence accuracy.
\newblock In \emph{International Conference on Machine Learning}, 2024.

\bibitem[Yuan et~al.(2021)Yuan, Ma, Zhang, Fang, and Wu]{yuan2021beyond}
Xiaoyong Yuan, Xiyao Ma, Lan Zhang, Yuguang Fang, and Dapeng Wu.
\newblock Beyond class-level privacy leakage: Breaking record-level privacy in
  federated learning.
\newblock \emph{IEEE Internet of Things Journal}, 9\penalty0 (4):\penalty0
  2555--2565, 2021.

\bibitem[Wang et~al.(2017)Wang, Blocki, Li, and Jha]{wang2017locally}
Tianhao Wang, Jeremiah Blocki, Ninghui Li, and Somesh Jha.
\newblock Locally differentially private protocols for frequency estimation.
\newblock In \emph{26th USENIX Security Symposium (USENIX Security 17)}, pages
  729--745, 2017.

\bibitem[Li et~al.(2023{\natexlab{a}})Li, Berrett, and Yu]{li2023robustness}
Mengchu Li, Thomas~B Berrett, and Yi~Yu.
\newblock On robustness and local differential privacy.
\newblock \emph{The Annals of Statistics}, 51\penalty0 (2):\penalty0 717--737,
  2023{\natexlab{a}}.

\bibitem[Wu et~al.(2022)Wu, Wu, Lyu, Qi, Huang, and Xie]{wu2022federated}
Chuhan Wu, Fangzhao Wu, Lingjuan Lyu, Tao Qi, Yongfeng Huang, and Xing Xie.
\newblock A federated graph neural network framework for privacy-preserving
  personalization.
\newblock \emph{Nature Communications}, 13\penalty0 (1):\penalty0 1--10, 2022.

\bibitem[Wang et~al.(2023{\natexlab{a}})Wang, Chen, Jiang, and
  Zhao]{wang2023ppefl}
Baocang Wang, Yange Chen, Hang Jiang, and Zhen Zhao.
\newblock Ppefl: Privacy-preserving edge federated learning with local
  differential privacy.
\newblock \emph{IEEE Internet of Things Journal}, 10\penalty0 (17):\penalty0
  15488--15500, 2023{\natexlab{a}}.

\bibitem[Batool et~al.(2024)Batool, Anjum, Khan, Izzo, Mazzocca, and
  Jeon]{batool2024secure}
Hajira Batool, Adeel Anjum, Abid Khan, Stefano Izzo, Carlo Mazzocca, and
  Gwanggil Jeon.
\newblock A secure and privacy preserved infrastructure for vanets based on
  federated learning with local differential privacy.
\newblock \emph{Information Sciences}, 652:\penalty0 119717, 2024.

\bibitem[Wang et~al.(2020{\natexlab{c}})Wang, Tong, and Shi]{wang2020federated}
Yansheng Wang, Yongxin Tong, and Dingyuan Shi.
\newblock Federated latent dirichlet allocation: A local differential privacy
  based framework.
\newblock In \emph{Proceedings of the AAAI Conference on Artificial
  Intelligence}, volume~34, pages 6283--6290, 2020{\natexlab{c}}.

\bibitem[Cheu et~al.(2019)Cheu, Smith, Ullman, Zeber, and
  Zhilyaev]{cheu2019distributed}
Albert Cheu, Adam Smith, Jonathan Ullman, David Zeber, and Maxim Zhilyaev.
\newblock Distributed differential privacy via shuffling.
\newblock In \emph{Proc. EUROCRYPT}, pages 375--403. Springer, 2019.

\bibitem[Chen et~al.(2022{\natexlab{a}})Chen, Choo, Kairouz, and
  Suresh]{chen2022fundamental}
Wei-Ning Chen, Christopher A~Choquette Choo, Peter Kairouz, and Ananda~Theertha
  Suresh.
\newblock The fundamental price of secure aggregation in differentially private
  federated learning.
\newblock In \emph{International Conference on Machine Learning (ICML)}, pages
  3056--3089. PMLR, 2022{\natexlab{a}}.

\bibitem[Agarwal et~al.(2018)Agarwal, Suresh, Yu, Kumar, and
  McMahan]{agarwal2018cpsgd}
Naman Agarwal, Ananda~Theertha Suresh, Felix Xinnan~X Yu, Sanjiv Kumar, and
  Brendan McMahan.
\newblock cpsgd: Communication-efficient and differentially-private distributed
  sgd.
\newblock In \emph{Advances in Neural Information Processing Systems
  (NeurIPS)}, volume~31, 2018.

\bibitem[Wei et~al.(2020)Wei, Li, Ding, Ma, Yang, Farokhi, Jin, Quek, and
  Poor]{wei2020federated}
Kang Wei, Jun Li, Ming Ding, Chuan Ma, Howard~H Yang, Farhad Farokhi, Shi Jin,
  Tony~QS Quek, and H~Vincent Poor.
\newblock Federated learning with differential privacy: Algorithms and
  performance analysis.
\newblock \emph{IEEE Transactions on Information Forensics and Security},
  15:\penalty0 3454--3469, 2020.

\bibitem[Cheng et~al.(2022)Cheng, Wang, Zhang, and
  Cheng]{cheng2022differentially}
Anda Cheng, Peisong Wang, Xi~Sheryl Zhang, and Jian Cheng.
\newblock Differentially private federated learning with local regularization
  and sparsification.
\newblock In \emph{Proceedings of the IEEE/CVF Conference on Computer Vision
  and Pattern Recognition (CVPR)}, pages 10122--10131, 2022.

\bibitem[Li et~al.(2023{\natexlab{b}})Li, Yang, Ren, Shi, and
  Zhao]{li2023multi}
Yanan Li, Shusen Yang, Xuebin Ren, Liang Shi, and Cong Zhao.
\newblock Multi-stage asynchronous federated learning with adaptive
  differential privacy.
\newblock \emph{IEEE Transactions on Pattern Analysis and Machine
  Intelligence}, 2023{\natexlab{b}}.

\bibitem[Chen et~al.(2020{\natexlab{b}})Chen, Wu, and
  Hong]{chen2020understanding}
Xiangyi Chen, Steven~Z Wu, and Mingyi Hong.
\newblock Understanding gradient clipping in private sgd: A geometric
  perspective.
\newblock In \emph{Advances in Neural Information Processing Systems
  (NeurIPS)}, pages 13773--13782, 2020{\natexlab{b}}.

\bibitem[Shi et~al.(2023{\natexlab{a}})Shi, Liu, Wei, Shen, Wang, and
  Tao]{shi2023make}
Yifan Shi, Yingqi Liu, Kang Wei, Li~Shen, Xueqian Wang, and Dacheng Tao.
\newblock Make landscape flatter in differentially private federated learning.
\newblock In \emph{Proceedings of the IEEE/CVF Conference on Computer Vision
  and Pattern Recognition (CVPR)}, pages 24552--24562, 2023{\natexlab{a}}.

\bibitem[Dai et~al.(2022)Dai, Shen, He, Tian, and Tao]{dai2022dispfl}
Rong Dai, Li~Shen, Fengxiang He, Xinmei Tian, and Dacheng Tao.
\newblock Dispfl: Towards communication-efficient personalized federated
  learning via decentralized sparse training.
\newblock In \emph{International Conference on Machine Learning (ICML)}, pages
  4587--4604. PMLR, 2022.

\bibitem[Hu et~al.(2021)Hu, Gong, and Guo]{hu2021federated}
Rui Hu, Yanmin Gong, and Yuanxiong Guo.
\newblock Federated learning with sparsification-amplified privacy and adaptive
  optimization.
\newblock In \emph{Proceedings of the Thirtieth International Joint Conference
  on Artificial Intelligence (IJCAI)}, 2021.

\bibitem[Li et~al.(2022{\natexlab{a}})Li, Wang, Chi, and
  Quek]{li2022differentially}
Yiwei Li, Shuai Wang, Chong-Yung Chi, and Tony~QS Quek.
\newblock Differentially private federated learning in edge networks: The
  perspective of noise reduction.
\newblock \emph{IEEE Network}, 36\penalty0 (5):\penalty0 167--172,
  2022{\natexlab{a}}.

\bibitem[Qi et~al.(2023)Qi, Wu, Wu, He, Huang, and Xie]{qi2023differentially}
Tao Qi, Fangzhao Wu, Chuhan Wu, Liang He, Yongfeng Huang, and Xing Xie.
\newblock Differentially private knowledge transfer for federated learning.
\newblock \emph{Nature Communications}, 14\penalty0 (1):\penalty0 3785, 2023.

\bibitem[Zhou et~al.(2024)Zhou, Li, Chen, Yang, Liang, and
  Zomaya]{zhou2024trustbcfl}
Sisi Zhou, Kuanching Li, Yuxiang Chen, Ce~Yang, Wei Liang, and Albert~Y Zomaya.
\newblock Trustbcfl: Mitigating data bias in iot through blockchain-enabled
  federated learning.
\newblock \emph{IEEE Internet of Things Journal}, 2024.

\bibitem[Qu et~al.(2021)Qu, Dai, Zhuang, Chen, Dong, Wu, and
  Guo]{qu2021decentralized}
Yuben Qu, Haipeng Dai, Yan Zhuang, Jiafa Chen, Chao Dong, Fan Wu, and Song Guo.
\newblock Decentralized federated learning for uav networks: Architecture,
  challenges, and opportunities.
\newblock \emph{IEEE Network}, 35\penalty0 (6):\penalty0 156--162, 2021.

\bibitem[Hashemi et~al.(2021)Hashemi, Acharya, Das, Vikalo, Sanghavi, and
  Dhillon]{hashemi2021benefits}
Abolfazl Hashemi, Anish Acharya, Rudrajit Das, Haris Vikalo, Sujay Sanghavi,
  and Inderjit Dhillon.
\newblock On the benefits of multiple gossip steps in communication-constrained
  decentralized federated learning.
\newblock \emph{IEEE Transactions on Parallel and Distributed Systems},
  33\penalty0 (11):\penalty0 2727--2739, 2021.

\bibitem[Gao et~al.(2023)Gao, Thai, and Wu]{gao2023decentralized}
Hongchang Gao, My~T Thai, and Jie Wu.
\newblock When decentralized optimization meets federated learning.
\newblock \emph{IEEE network}, 2023.

\bibitem[Sun et~al.(2023{\natexlab{b}})Sun, Li, and Wang]{9850408}
Tao Sun, Dongsheng Li, and Bao Wang.
\newblock Decentralized federated averaging.
\newblock \emph{IEEE Transactions on Pattern Analysis and Machine
  Intelligence}, 45\penalty0 (4):\penalty0 4289--4301, 2023{\natexlab{b}}.

\bibitem[Xu et~al.(2022)Xu, Zhang, and Wang]{9524471}
Jie Xu, Wei Zhang, and Fei Wang.
\newblock A(dp)$^2$sgd: Asynchronous decentralized parallel stochastic gradient
  descent with differential privacy.
\newblock \emph{IEEE Transactions on Pattern Analysis and Machine
  Intelligence}, 44\penalty0 (11):\penalty0 8036--8047, 2022.

\bibitem[Enrique Tomás Martínez~Beltrán and Celdrán(2023)]{10251949}
Pedro Miguel Sánchez Sánchez Sergio López Bernal Gérôme Bovet Manuel Gil
  Pérez Gregorio Martínez~Pérez Enrique Tomás Martínez~Beltrán, Mario
  Quiles~Pérez and Alberto~Huertas Celdrán.
\newblock Decentralized federated learning: fundamentals, state of the art,
  frameworks, trends, and challenges.
\newblock \emph{IEEE Communications surveys and tutorials}, 25\penalty0
  (4):\penalty0 2983--3013, 2023.

\bibitem[Ye et~al.(2022)Ye, Liang, and Li]{ye2022decentralized}
Hao Ye, Le~Liang, and Geoffrey~Ye Li.
\newblock Decentralized federated learning with unreliable communications.
\newblock \emph{IEEE journal of selected topics in signal processing},
  16\penalty0 (3):\penalty0 487--500, 2022.

\bibitem[Warnat-Herresthal et~al.(2021)Warnat-Herresthal, Schultze, Shastry,
  Manamohan, Mukherjee, Garg, Sarveswara, H{\"a}ndler, Pickkers, Aziz,
  et~al.]{warnat2021swarm}
Stefanie Warnat-Herresthal, Hartmut Schultze, Krishnaprasad~Lingadahalli
  Shastry, Sathyanarayanan Manamohan, Saikat Mukherjee, Vishesh Garg, Ravi
  Sarveswara, Kristian H{\"a}ndler, Peter Pickkers, N~Ahmad Aziz, et~al.
\newblock Swarm learning for decentralized and confidential clinical machine
  learning.
\newblock \emph{Nature}, 594\penalty0 (7862):\penalty0 265--270, 2021.

\bibitem[Wang et~al.(2021)Wang, Hu, Xiao, and Wu]{wang2021efficient}
Zhao Wang, Yifan Hu, Jun Xiao, and Chao Wu.
\newblock Efficient ring-topology decentralized federated learning with deep
  generative models for industrial artificial intelligent.
\newblock \emph{arXiv preprint arXiv:2104.08100}, 2021.

\bibitem[Bellet et~al.(2022)Bellet, Kermarrec, and Lavoie]{bellet2022d}
Aur{\'e}lien Bellet, Anne-Marie Kermarrec, and Erick Lavoie.
\newblock D-cliques: Compensating for data heterogeneity with topology in
  decentralized federated learning.
\newblock In \emph{2022 41st International Symposium on Reliable Distributed
  Systems (SRDS)}, pages 1--11. IEEE, 2022.

\bibitem[Vogels et~al.(2022)Vogels, Hendrikx, and Jaggi]{vogels2022beyond}
Thijs Vogels, Hadrien Hendrikx, and Martin Jaggi.
\newblock Beyond spectral gap: The role of the topology in decentralized
  learning.
\newblock In \emph{Advances in Neural Information Processing Systems
  (NeurIPS)}, volume~35, pages 15039--15050, 2022.

\bibitem[Mao et~al.(2020)Mao, Yuan, Hu, Gu, Sayed, and Yin]{mao2020walkman}
Xianghui Mao, Kun Yuan, Yubin Hu, Yuantao Gu, Ali~H Sayed, and Wotao Yin.
\newblock Walkman: A communication-efficient random-walk algorithm for
  decentralized optimization.
\newblock \emph{IEEE Transactions on Signal Processing}, 68:\penalty0
  2513--2528, 2020.

\bibitem[Mao et~al.(2018)Mao, Gu, and Yin]{mao2018walk}
Xianghui Mao, Yuantao Gu, and Wotao Yin.
\newblock Walk proximal gradient: An energy-efficient algorithm for consensus
  optimization.
\newblock \emph{IEEE Internet of Things Journal}, 6\penalty0 (2):\penalty0
  2048--2060, 2018.

\bibitem[Criado et~al.(2022)Criado, Casado, Iglesias, Regueiro, and
  Barro]{CRIADO2022263}
Marcos~F. Criado, Fernando~E. Casado, Roberto Iglesias, Carlos~V. Regueiro, and
  Senén Barro.
\newblock Non-iid data and continual learning processes in federated learning:
  A long road ahead.
\newblock \emph{Information Fusion}, 88:\penalty0 263--280, 2022.
\newblock ISSN 1566-2535.

\bibitem[Sheller et~al.(2020)Sheller, Edwards, Reina, Martin, Pati, Kotrotsou,
  Milchenko, Xu, Marcus, Colen, et~al.]{sheller2020federated}
Micah~J Sheller, Brandon Edwards, G~Anthony Reina, Jason Martin, Sarthak Pati,
  Aikaterini Kotrotsou, Mikhail Milchenko, Weilin Xu, Daniel Marcus, Rivka~R
  Colen, et~al.
\newblock Federated learning in medicine: facilitating multi-institutional
  collaborations without sharing patient data.
\newblock \emph{Scientific reports}, 10\penalty0 (1):\penalty0 12598, 2020.

\bibitem[Wang et~al.(2024{\natexlab{b}})Wang, Zhang, Su, and
  Zhu]{wang2024comprehensive}
Liyuan Wang, Xingxing Zhang, Hang Su, and Jun Zhu.
\newblock A comprehensive survey of continual learning: Theory, method and
  application.
\newblock \emph{IEEE Transactions on Pattern Analysis and Machine
  Intelligence}, 2024{\natexlab{b}}.

\bibitem[Wang et~al.(2023{\natexlab{b}})Wang, Zhang, Xu, Fu, Yang, and
  Du]{wang2023federated}
Zhe Wang, Yu~Zhang, Xinlei Xu, Zhiling Fu, Hai Yang, and Wenli Du.
\newblock Federated probability memory recall for federated continual learning.
\newblock \emph{Information Sciences}, 629:\penalty0 551--565,
  2023{\natexlab{b}}.

\bibitem[Dong et~al.(2023)Dong, Li, Cong, Sun, Zhang, and Van~Gool]{dong2023no}
Jiahua Dong, Hongliu Li, Yang Cong, Gan Sun, Yulun Zhang, and Luc Van~Gool.
\newblock No one left behind: Real-world federated class-incremental learning.
\newblock \emph{IEEE Transactions on Pattern Analysis and Machine
  Intelligence}, 2023.

\bibitem[Saldanha et~al.(2022)Saldanha, Quirke, West, James, Loughrey, Grabsch,
  Salto-Tellez, Alwers, Cifci, Ghaffari~Laleh, et~al.]{saldanha2022swarm}
Oliver~Lester Saldanha, Philip Quirke, Nicholas~P West, Jacqueline~A James,
  Maurice~B Loughrey, Heike~I Grabsch, Manuel Salto-Tellez, Elizabeth Alwers,
  Didem Cifci, Narmin Ghaffari~Laleh, et~al.
\newblock Swarm learning for decentralized artificial intelligence in cancer
  histopathology.
\newblock \emph{Nature medicine}, 28\penalty0 (6):\penalty0 1232--1239, 2022.

\bibitem[Nedic and Ozdaglar(2009)]{nedic2009distributed}
Angelia Nedic and Asuman Ozdaglar.
\newblock Distributed subgradient methods for multi-agent optimization.
\newblock \emph{IEEE Transactions on Automatic Control}, 54\penalty0
  (1):\penalty0 48--61, 2009.

\bibitem[Lopes and Sayed(2008)]{lopes2008diffusion}
Cassio~G Lopes and Ali~H Sayed.
\newblock Diffusion least-mean squares over adaptive networks: Formulation and
  performance analysis.
\newblock \emph{IEEE Transactions on Signal Processing}, 56\penalty0
  (7):\penalty0 3122--3136, 2008.

\bibitem[Koloskova et~al.(2020)Koloskova, Loizou, Boreiri, Jaggi, and
  Stich]{koloskova2020unified}
Anastasia Koloskova, Nicolas Loizou, Sadra Boreiri, Martin Jaggi, and Sebastian
  Stich.
\newblock A unified theory of decentralized sgd with changing topology and
  local updates.
\newblock In \emph{International Conference on Machine Learning (ICML)}, pages
  5381--5393. PMLR, 2020.

\bibitem[Zhang et~al.(2022{\natexlab{b}})Zhang, Fang, Liu, Yang, Liu, and
  Zhu]{zhang2022net}
Xin Zhang, Minghong Fang, Zhuqing Liu, Haibo Yang, Jia Liu, and Zhengyuan Zhu.
\newblock Net-fleet: Achieving linear convergence speedup for fully
  decentralized federated learning with heterogeneous data.
\newblock In \emph{Proceedings of the Twenty-Third International Symposium on
  Theory, Algorithmic Foundations, and Protocol Design for Mobile Networks and
  Mobile Computing (MobiHoc)}, pages 71--80, 2022{\natexlab{b}}.

\bibitem[Boyd et~al.(2004)Boyd, Diaconis, and Xiao]{boyd2004fastest}
Stephen Boyd, Persi Diaconis, and Lin Xiao.
\newblock Fastest mixing markov chain on a graph.
\newblock \emph{SIAM review}, 46\penalty0 (4):\penalty0 667--689, 2004.

\bibitem[Liang et~al.(2023)Liang, Peng, and
  Zhang]{liang2023gradienttrackinghighdimensional}
Jiadong Liang, Yang Peng, and Zhihua Zhang.
\newblock Gradient tracking for high dimensional federated optimization, 2023.
\newblock URL \url{https://arxiv.org/abs/2312.05590}.

\bibitem[Esfandiari et~al.(2021)Esfandiari, Tan, Jiang, Balu, Herron, Hegde,
  and Sarkar]{pmlr-v139-esfandiari21a}
Yasaman Esfandiari, Sin~Yong Tan, Zhanhong Jiang, Aditya Balu, Ethan Herron,
  Chinmay Hegde, and Soumik Sarkar.
\newblock Cross-gradient aggregation for decentralized learning from non-iid
  data.
\newblock In Marina Meila and Tong Zhang, editors, \emph{Proceedings of the
  38th International Conference on Machine Learning}, volume 139 of
  \emph{Proceedings of Machine Learning Research}, pages 3036--3046. PMLR,
  18--24 Jul 2021.

\bibitem[Kovalev et~al.(2021)Kovalev, Shulgin, Richtarik, Rogozin, and
  Gasnikov]{pmlr-v139-kovalev21a}
Dmitry Kovalev, Egor Shulgin, Peter Richtarik, Alexander~V Rogozin, and
  Alexander Gasnikov.
\newblock Adom: Accelerated decentralized optimization method for time-varying
  networks.
\newblock In Marina Meila and Tong Zhang, editors, \emph{International
  Conference on Machine Learning (ICML)}, volume 139 of \emph{Proceedings of
  Machine Learning Research}, pages 5784--5793. PMLR, 18--24 Jul 2021.

\bibitem[Ying et~al.(2021)Ying, Yuan, Chen, Hu, Pan, and
  Yin]{ying2021exponential}
Bicheng Ying, Kun Yuan, Yiming Chen, Hanbin Hu, Pan Pan, and Wotao Yin.
\newblock Exponential graph is provably efficient for decentralized deep
  training.
\newblock In \emph{Advances in Neural Information Processing Systems
  (NeurIPS)}, volume~34, pages 13975--13987, 2021.

\bibitem[Song et~al.(2022)Song, Li, Jin, Shi, Yan, Yin, and
  Yuan]{song2022communication}
Zhuoqing Song, Weijian Li, Kexin Jin, Lei Shi, Ming Yan, Wotao Yin, and Kun
  Yuan.
\newblock Communication-efficient topologies for decentralized learning with $
  o (1) $ consensus rate.
\newblock In \emph{Advances in Neural Information Processing Systems
  (NeurIPS)}, volume~35, pages 1073--1085, 2022.

\bibitem[Shi et~al.(2023{\natexlab{b}})Shi, Shen, Wei, Sun, Yuan, Wang, and
  Tao]{pmlr-v202-shi23d}
Yifan Shi, Li~Shen, Kang Wei, Yan Sun, Bo~Yuan, Xueqian Wang, and Dacheng Tao.
\newblock Improving the model consistency of decentralized federated learning.
\newblock In Andreas Krause, Emma Brunskill, Kyunghyun Cho, Barbara Engelhardt,
  Sivan Sabato, and Jonathan Scarlett, editors, \emph{Proceedings of the 40th
  International Conference on Machine Learning}, volume 202 of
  \emph{Proceedings of Machine Learning Research}, pages 31269--31291. PMLR,
  23--29 Jul 2023{\natexlab{b}}.

\bibitem[Zhang et~al.(2023{\natexlab{b}})Zhang, Liu, So, and
  Ling]{zhang2023variance}
Jiaojiao Zhang, Huikang Liu, Anthony Man-Cho So, and Qing Ling.
\newblock Variance-reduced stochastic quasi-newton methods for decentralized
  learning.
\newblock \emph{IEEE Transactions on Signal Processing}, 71:\penalty0 311--326,
  2023{\natexlab{b}}.

\bibitem[Xin et~al.(2020)Xin, Khan, and Kar]{xin2020variance}
Ran Xin, Usman~A Khan, and Soummya Kar.
\newblock Vari ance-reduced decentralized stochastic optimization with
  accelerated convergence.
\newblock \emph{IEEE Transactions on Signal Processing}, 68:\penalty0
  6255--6271, 2020.

\bibitem[Li et~al.(2022{\natexlab{b}})Li, Kailkhura, Goldhahn, Ray, and
  Varshney]{li2022robust}
Qunwei Li, Bhavya Kailkhura, Ryan Goldhahn, Priyadip Ray, and Pramod~K
  Varshney.
\newblock Robust decentralized learning using admm with unreliable agents.
\newblock \emph{IEEE Transactions on Signal Processing}, 70:\penalty0
  2743--2757, 2022{\natexlab{b}}.

\bibitem[Banerjee and Basu(2007)]{banerjee2007topic}
Arindam Banerjee and Sugato Basu.
\newblock Topic models over text streams: A study of batch and online
  unsupervised learning.
\newblock In \emph{Proceedings of the 2007 SIAM International Conference on
  Data Mining (SDM)}, pages 431--436. SIAM, 2007.

\bibitem[Hazan et~al.(2016)]{hazan2016introduction}
Elad Hazan et~al.
\newblock Introduction to online convex optimization.
\newblock \emph{Foundations and Trends{\textregistered} in Optimization},
  2\penalty0 (3-4):\penalty0 157--325, 2016.

\bibitem[Hoi et~al.(2021)Hoi, Sahoo, Lu, and Zhao]{hoi2021online}
Steven~CH Hoi, Doyen Sahoo, Jing Lu, and Peilin Zhao.
\newblock Online learning: A comprehensive survey.
\newblock \emph{Neurocomputing}, 459:\penalty0 249--289, 2021.

\bibitem[Dong et~al.(2022)Dong, Wang, Fang, Sun, Xu, Wang, and
  Zhu]{dong2022federated}
Jiahua Dong, Lixu Wang, Zhen Fang, Gan Sun, Shichao Xu, Xiao Wang, and Qi~Zhu.
\newblock Federated class-incremental learning.
\newblock In \emph{Proceedings of the IEEE/CVF Conference on Computer Vision
  and Pattern Recognition (CVPR)}, pages 10164--10173, 2022.

\bibitem[Gogineni et~al.(2022)Gogineni, Werner, Huang, and
  Kuh]{gogineniiot2022communication}
Vinay~Chakravarthi Gogineni, Stefan Werner, Yih-Fang Huang, and Anthony Kuh.
\newblock Communication-efficient online federated learning strategies for
  kernel regression.
\newblock \emph{IEEE Internet of Things Journal}, 10\penalty0 (5):\penalty0
  4531--4544, 2022.

\bibitem[Dekel et~al.(2008)Dekel, Shalev-Shwartz, and
  Singer]{dekel2008forgetron}
Ofer Dekel, Shai Shalev-Shwartz, and Yoram Singer.
\newblock The forgetron: A kernel-based perceptron on a budget.
\newblock \emph{SIAM Journal on Computing}, 37\penalty0 (5):\penalty0
  1342--1372, 2008.

\bibitem[Hoi et~al.(2013)Hoi, Jin, Zhao, and Yang]{hoi2013online}
Steven~CH Hoi, Rong Jin, Peilin Zhao, and Tianbao Yang.
\newblock Online multiple kernel classification.
\newblock \emph{Machine learning}, 90:\penalty0 289--316, 2013.

\bibitem[Yoon et~al.(2021)Yoon, Jeong, Lee, Yang, and Hwang]{yoon2021federated}
Jaehong Yoon, Wonyong Jeong, Giwoong Lee, Eunho Yang, and Sung~Ju Hwang.
\newblock Federated continual learning with weighted inter-client transfer.
\newblock In \emph{International Conference on Machine Learning (ICML)}, pages
  12073--12086. PMLR, 2021.

\bibitem[Babakniya et~al.(2024)Babakniya, Fabian, He, Soltanolkotabi, and
  Avestimehr]{babakniya2024data}
Sara Babakniya, Zalan Fabian, Chaoyang He, Mahdi Soltanolkotabi, and Salman
  Avestimehr.
\newblock A data-free approach to mitigate catastrophic forgetting in federated
  class incremental learning for vision tasks.
\newblock \emph{Advances in Neural Information Processing Systems (NeurIPS)},
  36, 2024.

\bibitem[Zhang et~al.(2023{\natexlab{c}})Zhang, Chen, Zhuang, and
  Lyu]{zhang2023target}
Jie Zhang, Chen Chen, Weiming Zhuang, and Lingjuan Lyu.
\newblock Target: Federated class-continual learning via exemplar-free
  distillation.
\newblock In \emph{Proceedings of the IEEE/CVF International Conference on
  Computer Vision (ICCV)}, pages 4782--4793, 2023{\natexlab{c}}.

\bibitem[Yuan et~al.(2023)Yuan, Ma, Su, and Wang]{yuan2023peer}
Liangqi Yuan, Yunsheng Ma, Lu~Su, and Ziran Wang.
\newblock Peer-to-peer federated continual learning for naturalistic driving
  action recognition.
\newblock In \emph{Proceedings of the IEEE/CVF Conference on Computer Vision
  and Pattern Recognition (CVPR)}, pages 5249--5258, 2023.

\bibitem[Ma et~al.(2022{\natexlab{a}})Ma, Xie, Wang, Chen, and
  Shou]{ma2022continual}
Yuhang Ma, Zhongle Xie, Jue Wang, Ke~Chen, and Lidan Shou.
\newblock Continual federated learning based on knowledge distillation.
\newblock In \emph{Proceedings of the 31st International Joint Conference on
  Artificial Intelligence (IJCAI)}, pages 2182--2188, 2022{\natexlab{a}}.

\bibitem[Shenaj et~al.(2023)Shenaj, Toldo, Rigon, and
  Zanuttigh]{shenaj2023asynchronous}
Donald Shenaj, Marco Toldo, Alberto Rigon, and Pietro Zanuttigh.
\newblock Asynchronous federated continual learning.
\newblock In \emph{Proceedings of the IEEE/CVF Conference on Computer Vision
  and Pattern Recognition (CVPR)}, pages 5054--5062, 2023.

\bibitem[Almanifi et~al.(2023)Almanifi, Chow, Tham, Chuah, and
  Kanesan]{almanifi2023communication}
Omair Rashed~Abdulwareth Almanifi, Chee-Onn Chow, Mau-Luen Tham, Joon~Huang
  Chuah, and Jeevan Kanesan.
\newblock Communication and computation efficiency in federated learning: A
  survey.
\newblock \emph{Internet of Things}, 22:\penalty0 100742, 2023.

\bibitem[Cho et~al.(2020)Cho, Wang, and Joshi]{cho2020client}
Yae~Jee Cho, Jianyu Wang, and Gauri Joshi.
\newblock Client selection in federated learning: Convergence analysis and
  power-of-choice selection strategies.
\newblock \emph{arXiv preprint arXiv:2010.01243}, 2020.

\bibitem[Miao et~al.(2023)Miao, Liu, Li, Li, Li, Choo, and
  Deng]{miao2023robust}
Yinbin Miao, Ziteng Liu, Xinghua Li, Meng Li, Hongwei Li, Kim-Kwang~Raymond
  Choo, and Robert~H Deng.
\newblock Robust asynchronous federated learning with time-weighted and stale
  model aggregation.
\newblock \emph{IEEE Transactions on Dependable and Secure Computing}, 2023.

\bibitem[Blanchard et~al.(2017)Blanchard, El~Mhamdi, Guerraoui, and
  Stainer]{blanchard2017machine}
Peva Blanchard, El~Mahdi El~Mhamdi, Rachid Guerraoui, and Julien Stainer.
\newblock Machine learning with adversaries: Byzantine tolerant gradient
  descent.
\newblock In \emph{Advances in Neural Information Processing Systems
  (NeurIPS)}, volume~30, 2017.

\bibitem[Khan et~al.(2021)Khan, Saad, Han, Hossain, and
  Hong]{khan2021federated}
Latif~U Khan, Walid Saad, Zhu Han, Ekram Hossain, and Choong~Seon Hong.
\newblock Federated learning for internet of things: Recent advances, taxonomy,
  and open challenges.
\newblock \emph{IEEE Communications Surveys \& Tutorials}, 23\penalty0
  (3):\penalty0 1759--1799, 2021.

\bibitem[Zhao et~al.(2023)Zhao, Mao, Liu, Song, Ouyang, Chen, and
  Ding]{zhao2023towards}
Zihao Zhao, Yuzhu Mao, Yang Liu, Linqi Song, Ye~Ouyang, Xinlei Chen, and Wenbo
  Ding.
\newblock Towards efficient communications in federated learning: A
  contemporary survey.
\newblock \emph{Journal of the Franklin Institute}, 360\penalty0 (12):\penalty0
  8669--8703, 2023.

\bibitem[Jiang et~al.(2022)Jiang, Wang, Li, and Yang]{jiang2022towards}
Zhifeng Jiang, Wei Wang, Bo~Li, and Qiang Yang.
\newblock Towards efficient synchronous federated training: A survey on system
  optimization strategies.
\newblock \emph{IEEE Transactions on Big Data}, 9\penalty0 (2):\penalty0
  437--454, 2022.

\bibitem[Luo et~al.(2022)Luo, Xiao, Wang, Huang, and
  Tassiulas]{luo2022tackling}
Bing Luo, Wenli Xiao, Shiqiang Wang, Jianwei Huang, and Leandros Tassiulas.
\newblock Tackling system and statistical heterogeneity for federated learning
  with adaptive client sampling.
\newblock In \emph{IEEE INFOCOM 2022-IEEE conference on computer communications
  (INFOCOM)}, pages 1739--1748. IEEE, 2022.

\bibitem[Fraboni et~al.(2021)Fraboni, Vidal, Kameni, and
  Lorenzi]{fraboni2021clustered}
Yann Fraboni, Richard Vidal, Laetitia Kameni, and Marco Lorenzi.
\newblock Clustered sampling: Low-variance and improved representativity for
  clients selection in federated learning.
\newblock In \emph{International Conference on Machine Learning (ICML)}, pages
  3407--3416. PMLR, 2021.

\bibitem[Reisizadeh et~al.(2020)Reisizadeh, Mokhtari, Hassani, Jadbabaie, and
  Pedarsani]{reisizadeh2020fedpaq}
Amirhossein Reisizadeh, Aryan Mokhtari, Hamed Hassani, Ali Jadbabaie, and
  Ramtin Pedarsani.
\newblock Fedpaq: A communication-efficient federated learning method with
  periodic averaging and quantization.
\newblock In \emph{International Conference on Artificial Intelligence and
  Statistics (AISTATS)}, pages 2021--2031. PMLR, 2020.

\bibitem[Han et~al.(2020)Han, Wang, and Leung]{han2020adaptive}
Pengchao Han, Shiqiang Wang, and Kin~K Leung.
\newblock Adaptive gradient sparsification for efficient federated learning: An
  online learning approach.
\newblock In \emph{2020 IEEE 40th International Conference on Distributed
  Computing Systems (ICDCS)}, pages 300--310. IEEE, 2020.

\bibitem[Eghlidi and Jaggi(2020)]{eghlidi2020sparse}
Negar~Foroutan Eghlidi and Martin Jaggi.
\newblock Sparse communication for training deep networks.
\newblock \emph{arXiv preprint arXiv:2009.09271}, 2020.

\bibitem[Sattler et~al.(2019)Sattler, Wiedemann, M{\"u}ller, and
  Samek]{sattler2019robust}
Felix Sattler, Simon Wiedemann, Klaus-Robert M{\"u}ller, and Wojciech Samek.
\newblock Robust and communication-efficient federated learning from non-iid
  data.
\newblock \emph{IEEE Transactions on neural networks and learning systems},
  31\penalty0 (9):\penalty0 3400--3413, 2019.

\bibitem[Liu et~al.(2020{\natexlab{c}})Liu, Cao, Yoshikawa, and
  Chen]{liu2020fedsel}
Ruixuan Liu, Yang Cao, Masatoshi Yoshikawa, and Hong Chen.
\newblock Fedsel: Federated sgd under local differential privacy with top-k
  dimension selection.
\newblock In \emph{Database Systems for Advanced Applications: 25th
  International Conference (DASFAA)}, pages 485--501. Springer,
  2020{\natexlab{c}}.

\bibitem[Lu et~al.(2023)Lu, Li, Liu, Guan, and Yang]{lu2023top}
Shiwei Lu, Ruihu Li, Wenbin Liu, Chaofeng Guan, and Xiaopeng Yang.
\newblock Top-k sparsification with secure aggregation for privacy-preserving
  federated learning.
\newblock \emph{Computers \& Security}, 124:\penalty0 102993, 2023.

\bibitem[Wangni et~al.(2018)Wangni, Wang, Liu, and Zhang]{wangni2018gradient}
Jianqiao Wangni, Jialei Wang, Ji~Liu, and Tong Zhang.
\newblock Gradient sparsification for communication-efficient distributed
  optimization.
\newblock In \emph{Advances in Neural Information Processing Systems
  (NeurIPS)}, volume~31, 2018.

\bibitem[Shi et~al.(2019)Shi, Chu, Cheung, and See]{shi2019understanding}
Shaohuai Shi, Xiaowen Chu, Ka~Chun Cheung, and Simon See.
\newblock Understanding top-k sparsification in distributed deep learning.
\newblock \emph{arXiv preprint arXiv:1911.08772}, 2019.

\bibitem[Shlezinger et~al.(2020)Shlezinger, Chen, Eldar, Poor, and
  Cui]{shlezinger2020uveqfed}
Nir Shlezinger, Mingzhe Chen, Yonina~C Eldar, H~Vincent Poor, and Shuguang Cui.
\newblock Uveqfed: Universal vector quantization for federated learning.
\newblock \emph{IEEE Transactions on Signal Processing}, 69:\penalty0 500--514,
  2020.

\bibitem[Alistarh et~al.(2017)Alistarh, Grubic, Li, Tomioka, and
  Vojnovic]{alistarh2017qsgd}
Dan Alistarh, Demjan Grubic, Jerry Li, Ryota Tomioka, and Milan Vojnovic.
\newblock Qsgd: Communication-efficient sgd via gradient quantization and
  encoding.
\newblock In \emph{Advances in Neural Information Processing Systems
  (NeurIPS)}, volume~30, 2017.

\bibitem[Suresh et~al.(2017)Suresh, Felix, Kumar, and
  McMahan]{suresh2017distributed}
Ananda~Theertha Suresh, X~Yu Felix, Sanjiv Kumar, and H~Brendan McMahan.
\newblock Distributed mean estimation with limited communication.
\newblock In \emph{International Conference on Machine Learning (ICML)}, pages
  3329--3337. PMLR, 2017.

\bibitem[Aji and Heafield(2017)]{aji2017sparse}
Alham~Fikri Aji and Kenneth Heafield.
\newblock Sparse communication for distributed gradient descent.
\newblock \emph{arXiv preprint arXiv:1704.05021}, 2017.

\bibitem[Zhou et~al.(2022)Zhou, Li, Ren, and Yang]{zhou2022towards}
Zihao Zhou, Yanan Li, Xuebin Ren, and Shusen Yang.
\newblock Towards efficient and stable k-asynchronous federated learning with
  unbounded stale gradients on non-iid data.
\newblock \emph{IEEE Transactions on Parallel and Distributed Systems},
  33\penalty0 (12):\penalty0 3291--3305, 2022.

\bibitem[Liu et~al.(2023)Liu, Chen, Lyu, Wu, Wu, and Chen]{liu2023byzantine}
Yuchen Liu, Chen Chen, Lingjuan Lyu, Fangzhao Wu, Sai Wu, and Gang Chen.
\newblock Byzantine-robust learning on heterogeneous data via gradient
  splitting.
\newblock In \emph{International Conference on Machine Learning (ICML)}, pages
  21404--21425. PMLR, 2023.

\bibitem[Zhu et~al.(2023)Zhu, Wang, Pang, Wang, Jiao, Song, and
  Jordan]{zhu2023byzantine}
Banghua Zhu, Lun Wang, Qi~Pang, Shuai Wang, Jiantao Jiao, Dawn Song, and
  Michael~I Jordan.
\newblock Byzantine-robust federated learning with optimal statistical rates.
\newblock In \emph{International Conference on Artificial Intelligence and
  Statistics (AISTATS)}, pages 3151--3178. PMLR, 2023.

\bibitem[Chen et~al.(2019)Chen, Sun, and Jin]{chen2019communication}
Yang Chen, Xiaoyan Sun, and Yaochu Jin.
\newblock Communication-efficient federated deep learning with layerwise
  asynchronous model update and temporally weighted aggregation.
\newblock \emph{IEEE Transactions on neural networks and learning systems},
  31\penalty0 (10):\penalty0 4229--4238, 2019.

\bibitem[Liu and Wright(2015)]{liu2015asynchronous}
Ji~Liu and Stephen~J Wright.
\newblock Asynchronous stochastic coordinate descent: Parallelism and
  convergence properties.
\newblock \emph{SIAM Journal on Optimization}, 25\penalty0 (1):\penalty0
  351--376, 2015.

\bibitem[Koloskova et~al.(2022)Koloskova, Stich, and
  Jaggi]{koloskova2022sharper}
Anastasiia Koloskova, Sebastian~U Stich, and Martin Jaggi.
\newblock Sharper convergence guarantees for asynchronous sgd for distributed
  and federated learning.
\newblock \emph{Advances in Neural Information Processing Systems},
  35:\penalty0 17202--17215, 2022.

\bibitem[Mitliagkas et~al.(2016)Mitliagkas, Zhang, Hadjis, and
  R{\'e}]{mitliagkas2016asynchrony}
Ioannis Mitliagkas, Ce~Zhang, Stefan Hadjis, and Christopher R{\'e}.
\newblock Asynchrony begets momentum, with an application to deep learning.
\newblock In \emph{2016 54th Annual Allerton Conference on Communication,
  Control, and Computing (Allerton)}, pages 997--1004. IEEE, 2016.

\bibitem[Dun et~al.(2023)Dun, Hipolito, Jermaine, Dimitriadis, and
  Kyrillidis]{dun2023efficient}
Chen Dun, Mirian Hipolito, Chris Jermaine, Dimitrios Dimitriadis, and
  Anastasios Kyrillidis.
\newblock Efficient and light-weight federated learning via asynchronous
  distributed dropout.
\newblock In \emph{International Conference on Artificial Intelligence and
  Statistics (AISTATS)}, pages 6630--6660. PMLR, 2023.

\bibitem[Du et~al.(2021)Du, Xu, Wu, and Tong]{du2021fairness}
Wei Du, Depeng Xu, Xintao Wu, and Hanghang Tong.
\newblock Fairness-aware agnostic federated learning.
\newblock In \emph{Proceedings of the 2021 SIAM International Conference on
  Data Mining (SDM)}, pages 181--189. SIAM, 2021.

\bibitem[Cui et~al.(2021)Cui, Pan, Liang, Zhang, and Wang]{cui2021addressing}
Sen Cui, Weishen Pan, Jian Liang, Changshui Zhang, and Fei Wang.
\newblock Addressing algorithmic disparity and performance inconsistency in
  federated learning.
\newblock In \emph{Advances in Neural Information Processing Systems
  (NeurIPS)}, volume~34, pages 26091--26102, 2021.

\bibitem[Yin et~al.(2018)Yin, Chen, Kannan, and Bartlett]{yin2018byzantine}
Dong Yin, Yudong Chen, Ramchandran Kannan, and Peter Bartlett.
\newblock Byzantine-robust distributed learning: Towards optimal statistical
  rates.
\newblock In \emph{International Conference on Machine Learning (ICML)}, pages
  5650--5659. PMLR, 2018.

\bibitem[Guerraoui et~al.(2018)Guerraoui, Rouault, et~al.]{guerraoui2018hidden}
Rachid Guerraoui, S{\'e}bastien Rouault, et~al.
\newblock The hidden vulnerability of distributed learning in byzantium.
\newblock In \emph{International Conference on Machine Learning (ICML)}, pages
  3521--3530. PMLR, 2018.

\bibitem[Farhadkhani et~al.(2022)Farhadkhani, Guerraoui, Gupta, Pinot, and
  Stephan]{farhadkhani2022byzantine}
Sadegh Farhadkhani, Rachid Guerraoui, Nirupam Gupta, Rafael Pinot, and John
  Stephan.
\newblock Byzantine machine learning made easy by resilient averaging of
  momentums.
\newblock In \emph{International Conference on Machine Learning (ICML)}, pages
  6246--6283. PMLR, 2022.

\bibitem[Karimireddy et~al.(2021{\natexlab{b}})Karimireddy, He, and
  Jaggi]{karimireddy2021byzantine}
Sai~Praneeth Karimireddy, Lie He, and Martin Jaggi.
\newblock Byzantine-robust learning on heterogeneous datasets via bucketing.
\newblock In \emph{International Conference on Learning Representations
  (ICLR)}, 2021{\natexlab{b}}.

\bibitem[Wu et~al.(2020{\natexlab{b}})Wu, Ling, Chen, and
  Giannakis]{wu2020federated}
Zhaoxian Wu, Qing Ling, Tianyi Chen, and Georgios~B Giannakis.
\newblock Federated variance-reduced stochastic gradient descent with
  robustness to byzantine attacks.
\newblock \emph{IEEE Transactions on Signal Processing}, 68:\penalty0
  4583--4596, 2020{\natexlab{b}}.

\bibitem[Gorbunov et~al.(2022)Gorbunov, Horv{\'a}th, Richt{\'a}rik, and
  Gidel]{gorbunovvariance}
Eduard Gorbunov, Samuel Horv{\'a}th, Peter Richt{\'a}rik, and Gauthier Gidel.
\newblock Variance reduction is an antidote to byzantines: Better rates, weaker
  assumptions and communication compression as a cherry on the top.
\newblock In \emph{The Eleventh International Conference on Learning
  Representations}, 2022.

\bibitem[Fedin and Gorbunov(2023)]{fedin2023byzantine}
Nikita Fedin and Eduard Gorbunov.
\newblock Byzantine-robust loopless stochastic variance-reduced gradient.
\newblock In \emph{International Conference on Mathematical Optimization Theory
  and Operations Research}, pages 39--53. Springer, 2023.

\bibitem[Su and Vaidya(2015)]{su2015fault}
Lili Su and Nitin~H Vaidya.
\newblock Fault-tolerant distributed optimization (part iv): Constrained
  optimization with arbitrary directed networks.
\newblock \emph{arXiv preprint arXiv:1511.01821}, 2015.

\bibitem[Fang et~al.(2022{\natexlab{b}})Fang, Yang, and Bajwa]{fang2022bridge}
Cheng Fang, Zhixiong Yang, and Waheed~U Bajwa.
\newblock Bridge: Byzantine-resilient decentralized gradient descent.
\newblock \emph{IEEE Transactions on Signal and Information Processing over
  Networks}, 8:\penalty0 610--626, 2022{\natexlab{b}}.

\bibitem[Yang and Bajwa(2019)]{yang2019byrdie}
Zhixiong Yang and Waheed~U Bajwa.
\newblock Byrdie: Byzantine-resilient distributed coordinate descent for
  decentralized learning.
\newblock \emph{IEEE Transactions on Signal and Information Processing over
  Networks}, 5\penalty0 (4):\penalty0 611--627, 2019.

\bibitem[Peng et~al.(2021)Peng, Li, and Ling]{peng2021byzantine}
Jie Peng, Weiyu Li, and Qing Ling.
\newblock Byzantine-robust decentralized stochastic optimization over static
  and time-varying networks.
\newblock \emph{Signal Processing}, 183:\penalty0 108020, 2021.

\bibitem[He et~al.(2022)He, Karimireddy, and Jaggi]{he2022byzantine}
Lie He, Sai~Praneeth Karimireddy, and Martin Jaggi.
\newblock Byzantine-robust decentralized learning via self-centered clipping.
\newblock \emph{arXiv preprint arXiv:2202.01545}, 2022.

\bibitem[Wu et~al.(2023)Wu, Chen, and Ling]{wu2023byzantine}
Zhaoxian Wu, Tianyi Chen, and Qing Ling.
\newblock Byzantine-resilient decentralized stochastic optimization with robust
  aggregation rules.
\newblock \emph{IEEE transactions on signal processing}, 2023.

\bibitem[Roth et~al.(2022)Roth, Cheng, Wen, Yang, Xu, Hsieh, Kersten, Harouni,
  Zhao, Lu, et~al.]{roth2022nvidia}
Holger~R Roth, Yan Cheng, Yuhong Wen, Isaac Yang, Ziyue Xu, Yuan-Ting Hsieh,
  Kristopher Kersten, Ahmed Harouni, Can Zhao, Kevin Lu, et~al.
\newblock Nvidia flare: Federated learning from simulation to real-world.
\newblock \emph{arXiv preprint arXiv:2210.13291}, 2022.

\bibitem[Posner et~al.(2021)Posner, Tseng, Aloqaily, and
  Jararweh]{posner2021federated}
Jason Posner, Lewis Tseng, Moayad Aloqaily, and Yaser Jararweh.
\newblock Federated learning in vehicular networks: Opportunities and
  solutions.
\newblock \emph{IEEE Network}, 35\penalty0 (2):\penalty0 152--159, 2021.

\bibitem[Savazzi et~al.(2021)Savazzi, Nicoli, Bennis, Kianoush, and
  Barbieri]{savazzi2021opportunities}
Stefano Savazzi, Monica Nicoli, Mehdi Bennis, Sanaz Kianoush, and Luca
  Barbieri.
\newblock Opportunities of federated learning in connected, cooperative, and
  automated industrial systems.
\newblock \emph{IEEE Communications Magazine}, 59\penalty0 (2):\penalty0
  16--21, 2021.

\bibitem[Wu et~al.(2020{\natexlab{c}})Wu, Cai, Xiao, Chen, and
  Ooi]{wu2020privacy}
Yuncheng Wu, Shaofeng Cai, Xiaokui Xiao, Gang Chen, and Beng~Chin Ooi.
\newblock Privacy preserving vertical federated learning for tree-based models.
\newblock \emph{arXiv preprint arXiv:2008.06170}, 2020{\natexlab{c}}.

\bibitem[Li et~al.(2024{\natexlab{a}})Li, Wang, Wang, Xia, Zhu, Chen, Fan,
  Cheng, and Lei]{li2024refer}
Wenjie Li, Zhongren Wang, Jinpeng Wang, Shu-Tao Xia, Jile Zhu, Mingjian Chen,
  Jiangke Fan, Jia Cheng, and Jun Lei.
\newblock Refer: Retrieval-enhanced vertical federated recommendation for full
  set user benefit.
\newblock In \emph{Proceedings of the 47th International ACM SIGIR Conference
  on Research and Development in Information Retrieval}, pages 1763--1773,
  2024{\natexlab{a}}.

\bibitem[Wu et~al.(2024)Wu, Li, Li, Ding, and Gao]{wu2024fedbiot}
Feijie Wu, Zitao Li, Yaliang Li, Bolin Ding, and Jing Gao.
\newblock Fedbiot: Llm local fine-tuning in federated learning without full
  model.
\newblock In \emph{Proceedings of the 30th ACM SIGKDD Conference on Knowledge
  Discovery and Data Mining}, pages 3345--3355, 2024.

\bibitem[Kang et~al.(2023)Kang, Fan, Gu, Fan, and Yang]{kang2023grounding}
Yan Kang, Tao Fan, Hanlin Gu, Lixin Fan, and Qiang Yang.
\newblock Grounding foundation models through federated transfer learning: A
  general framework.
\newblock \emph{arXiv preprint arXiv:2311.17431}, 2023.

\bibitem[Cheng et~al.(2021)Cheng, Chadha, and Duchi]{cheng2021fine}
Gary Cheng, Karan Chadha, and John Duchi.
\newblock Fine-tuning is fine in federated learning.
\newblock \emph{arXiv preprint arXiv:2108.07313}, 3, 2021.

\bibitem[Bergstra et~al.(2011)Bergstra, Bardenet, Bengio, and
  K{\'e}gl]{bergstra2011algorithms}
James Bergstra, R{\'e}mi Bardenet, Yoshua Bengio, and Bal{\'a}zs K{\'e}gl.
\newblock Algorithms for hyper-parameter optimization.
\newblock In \emph{Advances in Neural Information Processing Systems
  (NeurIPS)}, volume~24, 2011.

\bibitem[Bergstra and Bengio(2012)]{bergstra2012random}
James Bergstra and Yoshua Bengio.
\newblock Random search for hyper-parameter optimization.
\newblock \emph{Journal of machine learning research}, 13\penalty0 (2), 2012.

\bibitem[Yu and Zhu(2020)]{yu2020hyper}
Tong Yu and Hong Zhu.
\newblock Hyper-parameter optimization: A review of algorithms and
  applications.
\newblock \emph{arXiv preprint arXiv:2003.05689}, 2020.

\bibitem[White et~al.(2021)White, Neiswanger, and Savani]{white2021bananas}
Colin White, Willie Neiswanger, and Yash Savani.
\newblock Bananas: Bayesian optimization with neural architectures for neural
  architecture search.
\newblock In \emph{Proceedings of the AAAI Conference on Artificial
  Intelligence (AAAI)}, volume~35, pages 10293--10301, 2021.

\bibitem[Ren et~al.(2021)Ren, Xiao, Chang, Huang, Li, Chen, and
  Wang]{ren2021comprehensive}
Pengzhen Ren, Yun Xiao, Xiaojun Chang, Po-Yao Huang, Zhihui Li, Xiaojiang Chen,
  and Xin Wang.
\newblock A comprehensive survey of neural architecture search: Challenges and
  solutions.
\newblock \emph{ACM Computing Surveys (CSUR)}, 54\penalty0 (4):\penalty0 1--34,
  2021.

\bibitem[Zhang et~al.(2022{\natexlab{c}})Zhang, He, Gu, Gu, Deng, Huang, and
  Tao]{zhang2022bambi}
Qingsong Zhang, Fengxiang He, Jindong Gu, Bin Gu, Cheng Deng, Heng Huang, and
  Dacheng Tao.
\newblock Bambi: Vertical federated bilevel optimization with
  privacy-preserving and computation efficiency.
\newblock 2022{\natexlab{c}}.

\bibitem[Li et~al.(2024{\natexlab{b}})Li, Huang, and
  Huang]{li2024communication}
Junyi Li, Feihu Huang, and Heng Huang.
\newblock Communication-efficient federated bilevel optimization with global
  and local lower level problems.
\newblock In \emph{Advances in Neural Information Processing Systems
  (NeurIPS)}, volume~36, 2024{\natexlab{b}}.

\bibitem[Huang et~al.(2023)Huang, Zhang, and Ji]{huang2023achieving}
Minhui Huang, Dewei Zhang, and Kaiyi Ji.
\newblock Achieving linear speedup in non-iid federated bilevel learning.
\newblock In \emph{International Conference on Machine Learning (ICML)}, pages
  14039--14059. PMLR, 2023.

\bibitem[Tarzanagh et~al.(2022)Tarzanagh, Li, Thrampoulidis, and
  Oymak]{tarzanagh2022fednest}
Davoud~Ataee Tarzanagh, Mingchen Li, Christos Thrampoulidis, and Samet Oymak.
\newblock Fednest: Federated bilevel, minimax, and compositional optimization.
\newblock In \emph{International Conference on Machine Learning}, pages
  21146--21179. PMLR, 2022.

\bibitem[Yang et~al.(2024)Yang, Xiao, and Ji]{yang2024simfbo}
Yifan Yang, Peiyao Xiao, and Kaiyi Ji.
\newblock Simfbo: Towards simple, flexible and communication-efficient
  federated bilevel learning.
\newblock In \emph{Advances in Neural Information Processing Systems}, 2024.

\bibitem[Liu et~al.()Liu, Che, Zhou, Jin, Dai, Dou, and
  Valduriez]{doi:10.1137/1.9781611978032.95}
Ji~Liu, Tianshi Che, Yang Zhou, Ruoming Jin, Huaiyu Dai, Dejing Dou, and
  Patrick Valduriez.
\newblock Aedfl: Efficient asynchronous decentralized federated learning with
  heterogeneous devices.
\newblock In \emph{Proceedings of the 2024 SIAM International Conference on
  Data Mining (SDM)}, pages 833--841.

\bibitem[Keller(2020)]{keller2020mp}
Marcel Keller.
\newblock Mp-spdz: A versatile framework for multi-party computation.
\newblock In \emph{Proceedings of the 2020 ACM SIGSAC conference on computer
  and communications security (CCS)}, pages 1575--1590, 2020.

\bibitem[Kadhe et~al.(2020)Kadhe, Rajaraman, Koyluoglu, and
  Ramchandran]{kadhe2020fastsecagg}
Swanand Kadhe, Nived Rajaraman, O~Ozan Koyluoglu, and Kannan Ramchandran.
\newblock Fastsecagg: Scalable secure aggregation for privacy-preserving
  federated learning.
\newblock \emph{arXiv preprint arXiv:2009.11248}, 2020.

\bibitem[Acar et~al.(2018)Acar, Aksu, Uluagac, and Conti]{acar2018survey}
Abbas Acar, Hidayet Aksu, A~Selcuk Uluagac, and Mauro Conti.
\newblock A survey on homomorphic encryption schemes: Theory and
  implementation.
\newblock \emph{ACM Computing Surveys (Csur)}, 51\penalty0 (4):\penalty0 1--35,
  2018.

\bibitem[Ma et~al.(2022{\natexlab{b}})Ma, Naas, Sigg, and Lyu]{ma2022privacy}
Jing Ma, Si-Ahmed Naas, Stephan Sigg, and Xixiang Lyu.
\newblock Privacy-preserving federated learning based on multi-key homomorphic
  encryption.
\newblock \emph{International Journal of Intelligent Systems}, 37\penalty0
  (9):\penalty0 5880--5901, 2022{\natexlab{b}}.

\bibitem[Zhang et~al.(2020)Zhang, Li, Xia, Wang, Yan, and
  Liu]{zhang2020batchcrypt}
Chengliang Zhang, Suyi Li, Junzhe Xia, Wei Wang, Feng Yan, and Yang Liu.
\newblock $\{$BatchCrypt$\}$: Efficient homomorphic encryption for
  $\{$Cross-Silo$\}$ federated learning.
\newblock In \emph{2020 USENIX annual technical conference (USENIX ATC)}, pages
  493--506, 2020.

\bibitem[Bonawitz et~al.(2017)Bonawitz, Ivanov, Kreuter, Marcedone, McMahan,
  Patel, Ramage, Segal, and Seth]{bonawitz2017practical}
Keith Bonawitz, Vladimir Ivanov, Ben Kreuter, Antonio Marcedone, H~Brendan
  McMahan, Sarvar Patel, Daniel Ramage, Aaron Segal, and Karn Seth.
\newblock Practical secure aggregation for privacy-preserving machine learning.
\newblock In \emph{Proceedings of the 2017 ACM SIGSAC Conference on Computer
  and Communications Security (CCS)}, pages 1175--1191, 2017.

\bibitem[Chen et~al.(2022{\natexlab{b}})Chen, Ozgur, and
  Kairouz]{chen2022poisson}
Wei-Ning Chen, Ayfer Ozgur, and Peter Kairouz.
\newblock The poisson binomial mechanism for unbiased federated learning with
  secure aggregation.
\newblock In \emph{International Conference on Machine Learning}, pages
  3490--3506. PMLR, 2022{\natexlab{b}}.

\end{thebibliography}

\end{document}